\documentclass[10pt,twocolumn,letterpaper]{article}

\usepackage[pagenumbers]{cvpr}

\usepackage{xcolor}
\usepackage{placeins}
\usepackage{colortbl}
\usepackage{pifont}

\usepackage{tikz}
\usetikzlibrary{arrows.meta,positioning,fit,calc}

\usepackage{xcolor}

\definecolor{LightCyan}{rgb}{0.88,1,1}
\newcommand{\tts}{\textsc{TTS}}

\usepackage{multirow}

\definecolor{cvprblue}{rgb}{0.21,0.49,0.74}
\usepackage[pagebackref,breaklinks,colorlinks,allcolors=cvprblue]{hyperref}

\def\confName{CVPR}
\def\confYear{2027}

\title{A Test-Time Solver for Uncertainty-Aware 4D Novel View Synthesis}

\author{
Miaowei Wang$^{1}$ \quad
Zhi Cao$^{2}$ \quad
Jingwei Xu$^{3}$ \quad
Oisin Mac Aodha$^{1}$ \quad
Amir Vaxman$^{1}$\\[2pt]
$^{1}$University of Edinburgh \quad
$^{2}$Stanford University \quad
$^{3}$NovaCore\\[2pt]
{\small Project page: \url{https://wangmiaowei.github.io/TTS4D/}}
}

\begin{document}
\twocolumn[{%
\renewcommand\twocolumn[1][]{#1}%
\maketitle
\begin{center}
    \centering
    \captionsetup{type=figure}
    \vspace{-9mm}
    \includegraphics[width=0.92\textwidth]{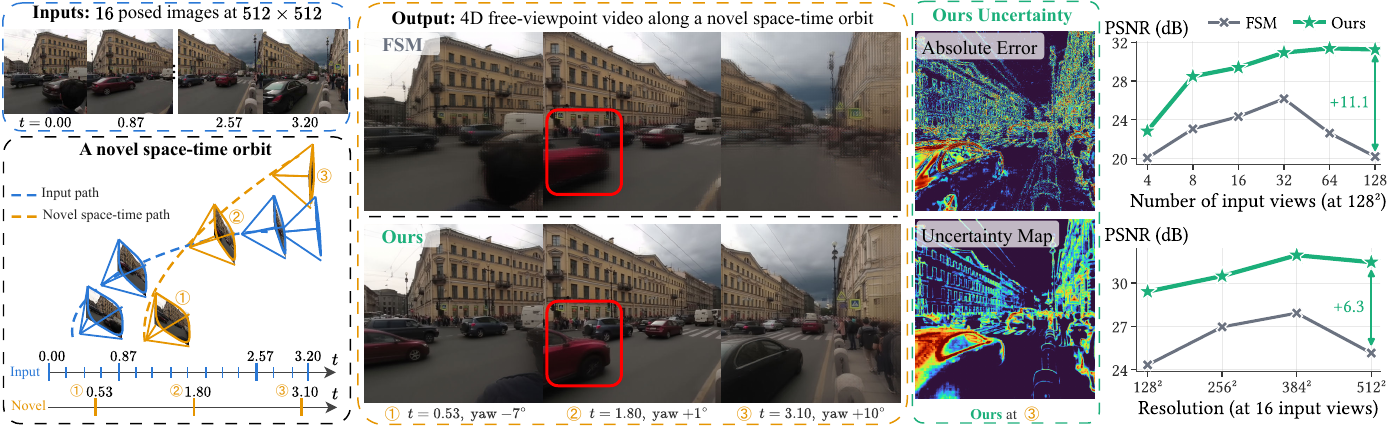}
    \vspace{-10pt}
\captionof{figure}{Our feed-forward 4D novel view synthesis model, built on our test-time solver layer, renders a dynamic scene from 16 posed input images along a novel space-time camera orbit, at $512\times512$ on one H20 GPU at 22\,FPS. The state-of-the-art FSM~\cite{fsm} blurs the moving car (red box), whereas ours keeps it sharp and also outputs an uncertainty map that correlates with the error. As the number of input views and the resolution increase, our results are robust while FSM degrades sharply. See our project page for the full motion dynamics.}

    \label{fig:teaser}
\end{center}
\vspace{-2mm}
}]

\begin{abstract}
4D novel view synthesis (NVS) methods require many input frames to faithfully reconstruct long, complex, and dynamic scenes.   
However, current Transformer-based methods are bottlenecked by the quadratic scaling of softmax attention. 
Test-time training (TTT) attempts to address this limitation by compressing tokens into fast weights at linear cost, but introduces a dependency on token order that can affect performance, as well as the need for gradient-based optimization at inference time.
We propose a novel Test-time solver (TTS) layer that solves the same fitting via a closed-form expression, resulting in an order-invariant approach that trains nearly $2\times$ faster than comparable TTT models. 
Our TTS solution also naturally allows us to represent uncertainty in the model's predictions, a desirable capability given that NVS is highly under-constrained as parts of a scene may not be visible in the input images. 
By combining an epistemic term from our solver's posterior with a learned aleatoric term, we can generate an uncertainty map in a feed-forward manner. 
Our approach sets a new state of the art on multiple 4D NVS benchmarks across different scene types, image resolutions, and numbers of input views. 
In addition, our uncertainty maps correlate with parts of the image where the model's predictions are inaccurate, providing an important step toward reliable 4D NVS. 
\end{abstract}
\section{Introduction}
\label{sec:intro}
\vspace{-5pt}
The goal of 4D novel view synthesis (NVS) is to render a \emph{dynamic} scene
from new camera viewpoints at any point in time, with applications in virtual
reality~\cite{broxton2020immersive}, free-viewpoint video~\cite{lin2023high},
robotics~\cite{zhou2023nerf} and driving simulation~\cite{yan2024street}.
Optimization-based
methods~\cite{mildenhall2020nerf,kerbl2023gaussian,pumarola2021dnerf,wu20244dgs}
achieve high-quality results but need to fit each scene from scratch, taking
minutes to hours. Feed-forward models, trained on many
scenes~\cite{hong2024lrm,charatan2024pixelsplat,jin2025lvsm,4dgt,movies},
instead predict novel views in a single network pass. Earlier models were
conditioned on only two to four input
images~\cite{charatan2024pixelsplat,chen2024mvsplat,gslrm,jin2025lvsm}, too
few to cover a large scene. 
Moreover, in dynamic scenes, a moving object may be
seen from each viewpoint only briefly, thus an accurate novel view needs evidence
from many frames, necessitating NVS models that condition on tens to hundreds
of input views~\cite{longlrm,movies,zhang2026test,kang2026multi,fsm}.

However, processing many input views is expensive. Most feed-forward models are Transformer-based~\cite{vaswani2017attention} and split each view into multiple patch tokens. 
Naïve softmax attention~\cite{bahdanau2015neural} compares all pairs of tokens in every layer, and its cost grows quadratically with the number of views and image resolution. To reduce this cost, linear
attention~\cite{katharopoulos2020transformers} and its
successors~\cite{ahn2024linear,han2023flatten,schlag2021linear,gu2023mamba,yang2024parallelizing}
attempt to compress all inputs into a fixed-size \emph{memory}.
Test-time training (TTT)~\cite{ttt} makes this memory the weights of a small
neural network in each layer. Unlike ordinary weights, which are fixed after
training, these \emph{fast weights}~\cite{schlag2021linear,ba2016using} are
fit anew to every input sequence at test time.
Input tokens are \emph{written} into the memory by training the network at test time to map their keys to their values. 
Tokens of the target rendered view \emph{read} the memory by passing their query vectors through the network. 
Writing is thus linear in the number of input views, and reading is constant time per token.  
Inspired by \textsc{LaCT}~\cite{zhang2026test}, \textsc{FSM}~\cite{fsm} recently introduced a state-of-the-art (SOTA) 4D NVS model.

TTT's weakness is in the memory writing. Each chunk of input tokens is
written with a single gradient step, which fits the chunk only roughly. As a
result, reading a token back returns not only its own value but blended with
values of other, even unrelated, tokens (\cref{fig: relative_weihts}). This
interference grows with every token written, \ie, with more views and at
higher resolution, and blurs the renderings (\cref{fig:teaser}). Writing
chunk by chunk also overwrites earlier chunks and makes the memory depend on
the view order. \textsc{FSM}~\cite{fsm} reduces this forgetting with elastic weight
consolidation~\cite{kirkpatrick2017overcoming}, but this does not remove the
interference.

Our key observation is that with a single fast-weight matrix, this fitting problem can be solved exactly in closed form at test time instead of approximately by one gradient step. 
We call this a \emph{test-time solver} (TTS) and build an NVS model on it. 
The solver fits all input tokens jointly and discounts what their keys share, so reading a token returns mainly its own value and those of similar tokens, rather than a weighted sum of all written values. 
Regularizing the solution toward a learned prior helps fill in regions the inputs miss, especially with sparse views. 
A 2D-time RoPE~\cite{su2024roformer} in TTS links tokens close in image position or
time, so neighboring patches blend smoothly. 
TTS depends on the views only through sums, so their order does not matter, and both the solve and its gradient reduce to dense matrix products, making each training step nearly $2\times$ faster than \textsc{FSM}'s. 

4D NVS is a very challenging under-constrained task due to occlusion, unseen regions, ambiguous geometry, and motion between observed input images.
Thus, knowing which generated pixels are unreliable is highly valuable in applications; \eg,
a driving system can distrust those pixels~\cite{mcallister2017concrete}, a
robot can capture the missing views~\cite{pan2022activenerf,jiang2024fisherrf},
or a generative model can fill them in~\cite{tan2025extrags,bose2025uarscenes}.
Unlike time-consuming per-scene methods~\cite{martinbrualla2021nerfw,shen2021snerf,goli2024bayesrays,jiang2024fisherrf},
our model provides this at low cost. 
Interpreting TTS as Bayesian linear regression~\cite{bishop2006prml} yields a posterior that quantifies the uncertainty in each part of the target view with regard to the inputs. We output this as an uncertainty map, computed in the same forward pass as the rendering. 

Trained on many diverse datasets, our model outperforms existing baselines on multiple NVS benchmarks under different image resolutions and numbers of input views. 
Furthermore, our predicted uncertainty map correlates with rendering errors, flagging unreliable regions. 

In summary, our contributions are:
\begin{itemize}
\item a test-time solver (TTS), whose fast weights are solved in closed form
instead of gradient steps;
\item a feed-forward 4D NVS model built on TTS that achieves SOTA
results on multiple 4D benchmarks;
\item the first feed-forward uncertainty map for 4D NVS.
\end{itemize}
\begin{figure*}[!ht]
\centering
\includegraphics[width=0.95\linewidth]{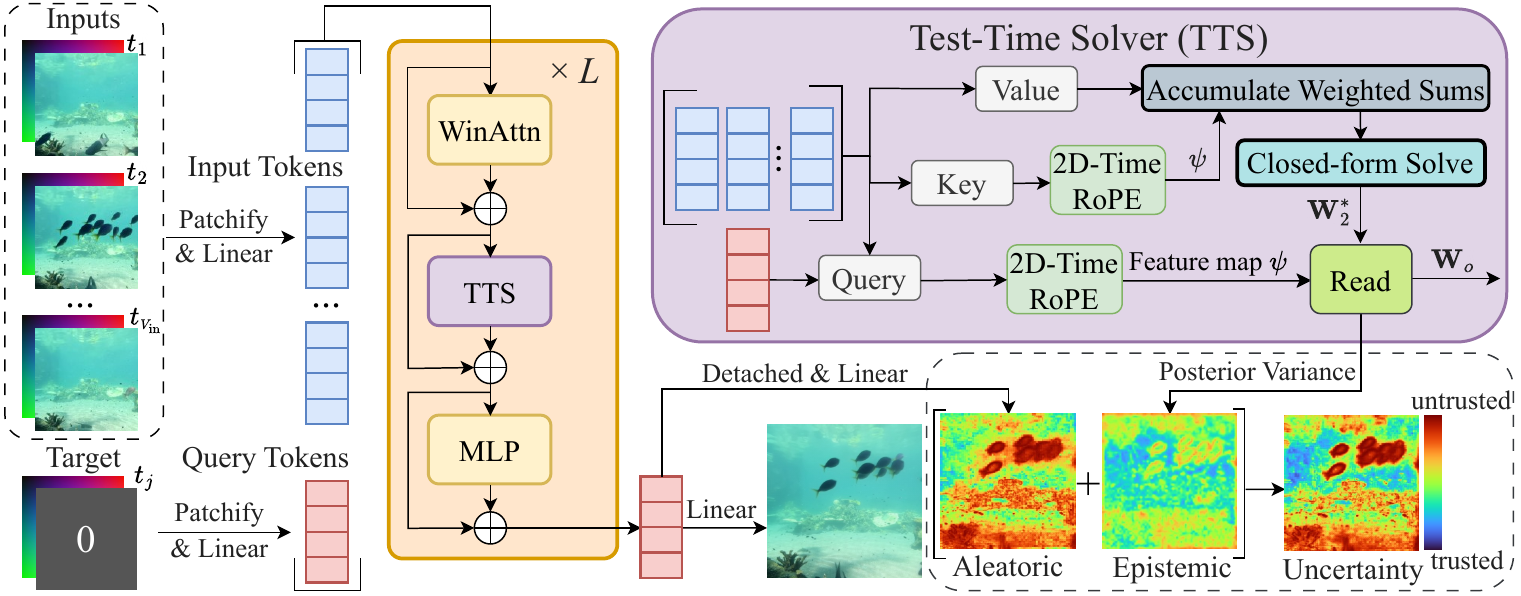}
\vspace{-3mm}
\caption{\textbf{Model overview}. From posed input images captured at different
times, our NVS model renders target views at new camera poses and times, each with
an uncertainty map. 
Views exchange information only through our test-time
solver (TTS), which accumulates all input tokens into weighted sums of fixed
size and solves for the single fast weight $\mathbf W_2$ of
each TTS layer via a closed-form expression.
}
\vspace{-5mm}
\label{fig: pipeline}
\end{figure*}

\section{Related Work}
\label{sec:related}
\noindent\textbf{Feed-forward 4D NVS.}
Instead of performing per-scene optimization~\cite{mildenhall2020nerf,kerbl2023gaussian},
feed-forward NVS models make predictions in one network pass, either via
explicit 3D representations~\cite{hong2024lrm,charatan2024pixelsplat,chen2024mvsplat,gslrm,wang2025decoupledgaussian}
or directly as images~\cite{jin2025lvsm,jiang2025rayzer,szymanowicz2026lagernvs}.
Hybrid or fast-weight architectures scale them to many views~\cite{longlrm,zhang2026test,wang2026tttlrm}. 
Dynamic scenes add time and require many more context frames as input. 
Most 4D models predict Gaussians that change
over time~\cite{l4gm,yang2025storm,liang2025bullet,lin2025dgs,ma2025dlrm,wang2025canfields}.
\textsc{4DGT}~\cite{4dgt} learns 4D Gaussians with limited lifespans from posed
monocular videos and covers long videos with a rolling window.
\textsc{MoVieS}~\cite{movies} predicts one Gaussian per pixel and uses a motion head that
moves it to any queried time. \textsc{NeoVerse}~\cite{neoverse} reconstructs 4D
Gaussians with a backbone built on \textsc{VGGT}~\cite{wang2025vggt} and pairs them
with a video diffusion model. \textsc{C4G}~\cite{c4g} proposes a compact set of time-conditioned Gaussian
tokens, shared by the whole video, and requires no cameras. \textsc{FL4N}~\cite{fl4n}
avoids explicit primitives and learns dynamics as a latent residual on a
pretrained static model~\cite{szymanowicz2026lagernvs}. Video diffusion
models~\cite{wu2025cat4d,vanhoorick2024gcd,anyview,wang2026motionphysics} also render dynamic scenes from new viewpoints, but their stochastic, multi-step sampling is slow and can contradict the inputs. Closest to us, \textsc{FSM}~\cite{fsm} extends \textsc{LaCT}~\cite{zhang2026test} to 4D, but its single gradient step per chunk results in interference that increases with frames and resolution.

\vspace{0.5mm}
\noindent\textbf{Linear attention.}
To reduce the cost of softmax attention, some methods restrict each token to a local
window~\cite{zhang2026gamma,kang2026multi} or compress the tokens into a
few representative tokens~\cite{Wang2025CLiFT,veicht2026zipsplat,wen2024efficient,szymanowicz2026lagernvs}.
Linear attention~\cite{katharopoulos2020transformers} and its
variants~\cite{schlag2021linear,gu2023mamba} instead sum key--value outer
products into a constant-size state, so non-orthogonal keys interfere at
read-out. Such layers can be seen as a key-to-value regression problem that is solved at test
time~\cite{wang2025ttr}. Most approximate it with a few gradient
steps~\cite{yang2024parallelizing,gateddeltanet,deltaproduct,longhorn,ttt,behrouz2026titans,atlas,zhang2026test,fsm},
which do not reach the optimum~\cite{vonoswald2023gd,ahn2023preconditioned}
and, when applied chunk by chunk, depend on how the views are partitioned and
ordered. 
Others solve it exactly, \eg, for language modeling, image classification, or few-shot learning~\cite{mesa,gatedkalmanet,intention}. 
Closest to us, \textsc{MesaNet}~\cite{mesanet} solves a regularized least-squares
problem with forgetting gates at every token, for causal language modeling.
As input views only write and target views only read, we solve the regression
over the whole input set in closed form. 
To our knowledge, ours is the first solved memory for feed-forward NVS.

\vspace{0.5mm}
\noindent\textbf{Uncertainty in NVS.}
NVS methods can output a plausible color for every pixel, even where no input
view observes that part of the scene, so failures can go unnoticed unless
uncertainty is also reported. Most methods estimate it per scene, during or
after optimization~\cite{martinbrualla2021nerfw,shen2021snerf,goli2024bayesrays,jiang2024fisherrf,var3dgs,lyu2024manifold,galappaththige2026predictive},
at extra cost for every scene. The few that handle dynamic scenes use uncertainty to
regularize poorly observed regions~\cite{kim2024wild4dgs,guo2026usplat4d}
rather than reporting it. Feed-forward models predict it with an extra head, as a
per-pixel color variance~\cite{jin2023neunbv} or confidence in  predicted
geometry~\cite{wang2024dust3r,wang2025vggt}, but only for static scenes. To
our knowledge, ours is the first feed-forward per-pixel uncertainty map for 4D~NVS.
\section{Methodology}
\label{sec:model}

\subsection{Problem Statement}
\label{sec:task} 

Given a set of input views of a dynamic scene (\eg, people walking, moving vehicles, \etc), our goal is to synthesize images of the scene from
new camera viewpoints and at new times. Specifically, we take $V_{\mathrm{in}}$
posed images $\{\mathcal I_i\in\mathbb R^{H\times W\times3}\}_{i=1}^{V_{\mathrm{in}}}$
of the scene as input, each with camera intrinsics
$\mathbf K_i\in\mathbb R^{3\times3}$, pose $\mathbf E_i\in\mathrm{SE}(3)$, and
timestamp $t_i\in\mathbb R$. The images can be from any viewpoint at any time,
in no particular order. For a target view with camera
$(\mathbf K_j,\mathbf E_j)$ at time $t_j$, we aim to predict in one forward
pass the image $\mathcal I_j$ and a per-pixel uncertainty
map~$\mathcal U_j\in\mathbb R^{H\times W}$. 

\subsection{Preliminaries: Attention and Fast weights}
\label{sec:ttt}

\noindent\textbf{Softmax attention.}
Consider a sequence of $N$ tokens $\{\mathbf x_n\in\mathbb R^d\}_{n=1}^N$,
obtained by encoding the input views. 
In a Transformer layer, tokens exchange information by content-based lookup~\cite{vaswani2017attention}. 
Each token $\mathbf x_n$ is projected to a query $\mathbf q_n$, a key $\mathbf k_n$ and a value $\mathbf v_n$ in
$\mathbb R^d$. A query is compared with every key and retrieves the values of
the keys it matches,
$\mathbf o_m=\sum_{n=1}^{N}\operatorname{softmax}_n\big(\mathbf q_m^{\top}\mathbf k_n/\sqrt d\big)\,\mathbf v_n$.
All pairs of $\{(\mathbf k_n,\mathbf v_n)\}_{n=1}^N$ can be viewed as a \emph{memory} that
grows linearly with $N$, and computing $\mathbf o_m$ for all $N$ tokens costs
$O(N^2)$, a bottleneck for long sequences~\cite{zhang2026test,fsm}.

\vspace{0.5mm}
\noindent\textbf{Fast weights.}
Test-time training (TTT)~\cite{ttt} replaces this memory with \emph{fast} weights~\cite{schmidhuber1992learning,ba2016using,schlag2021linear}. 
Unlike traditional (\ie \emph{slow}) weights, which are learned only in training and fixed at inference, fast weights are also adapted during inference. 
TTT \emph{writes} the inputs into memory by treating the key--value pairs as training data and fitting a map $f_\theta:\mathbb R^{d}\to\mathbb R^{d}$ with $f_\theta(\hat{\mathbf k}_n)\approx\mathbf v_n$, where $\hat{\mathbf k}_n=\mathbf k_n/\|\mathbf k_n\|$ and likewise for $\hat{\mathbf q}_m$. 
The memory is then the fast weights $\theta$, of fixed size and initialized from learned $\theta^{\mathrm{init}}$. 
A query \emph{reads} it as in attention, but is passed through $f_\theta$ instead of being compared with every key, so a read $f_\theta(\hat{\mathbf q}_m)$ costs the same for any $N$. 
In LaCT~\cite{zhang2026test} and FSM~\cite{fsm}, $f_\theta$ is a SwiGLU MLP~\cite{shazeer2020glu}:
\vspace{-1mm}
\begin{equation}
  f_\theta(\mathbf z)=\mathbf W_2\,\phi(\mathbf z),\quad
  \phi(\mathbf z)=\operatorname{SiLU}(\mathbf W_1\mathbf z)\odot\mathbf W_3\mathbf z,
  \label{eq:swiglu}
\vspace{-1mm}
\end{equation}
with fast weights $\theta=\{\mathbf W_1,\mathbf W_3\in\mathbb R^{h\times d},\
\mathbf W_2\in\mathbb R^{d\times h}\}$, where the input $\mathbf z$ is
$\hat{\mathbf k}_n$ when writing and $\hat{\mathbf q}_m$ when reading.

For each chunk $c$ of thousands of tokens, FSM updates the fast weights by taking one step $\theta\leftarrow\operatorname{Norm}\big(\theta-\operatorname{Muon}(\nabla_\theta\mathcal L_{\mathrm{FSM}})\big)$ on the dot-product loss $\mathcal L_{\mathrm{FSM}}=-\sum_{n\in c}a_n\mathbf v_n^{\top}f_\theta(\hat{\mathbf k}_n)$, where $a_n>0$ are learned per-token weights. 
$\operatorname{Muon}$~\cite{jordan2024muon} orthogonalizes the gradient and $\operatorname{Norm}$ restores the weight norms. 
FSM~\cite{fsm} also uses elastic weight consolidation~\cite{kirkpatrick2017overcoming}, which, after each chunk, pulls the weights that earlier chunks rely on toward a moving average of their past values.

\begin{figure}[!t]
\centering
\includegraphics[width=0.95\linewidth]{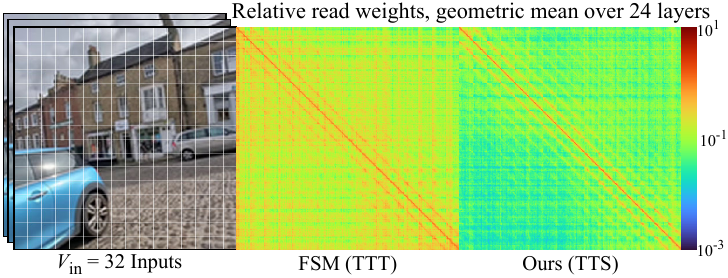}
\vspace{-3mm}
\caption{\textbf{Relative read weights.} Weight of token $n$'s value
(column) in token $m$'s read (row), relative to $m$'s own, in the first view
(computed by automatic differentiation). FSM~\cite{fsm} (TTT with
$\operatorname{Muon}$, $\operatorname{Norm}$, and consolidation) spreads each
read over many tokens, whereas TTS (w/o.~RoPE) returns mainly each token and its spatial
neighbors (diagonal bands), suppressing interference.}
\vspace{-5mm}
\label{fig: relative_weihts}
\end{figure}

\vspace{0.5mm}
\noindent\textbf{Write interference.}
Reading a written key should return its value, $f_\theta(\hat{\mathbf k}_m)\approx\mathbf v_m$, which FSM's one-step write does not achieve. 
Consider a plain gradient step on $\mathbf W_2$ alone, without $\operatorname{Muon}$, $\operatorname{Norm}$ or updates of $\mathbf W_1,\mathbf W_3$. 
With $\phi_n:=\phi(\hat{\mathbf k}_n)$, each chunk then adds $\sum_{n\in c}a_n\mathbf v_n\phi_n^{\top}$ to $\mathbf W_2$, as in linear attention~\cite{katharopoulos2020transformers,schlag2021linear}. 
After all $N$ tokens are written with unit step size, the read of $\hat{\mathbf k}_m$ returns: 
\vspace{-1mm}
\begin{equation}
  \mathbf W_2^{\mathrm{init}}\phi_m+a_m\|\phi_m\|^2\,\mathbf v_m
  +\textstyle\sum_{n\neq m}a_n\langle\phi_n,\phi_m\rangle\,\mathbf v_n .
  \label{eq:readback}
\vspace{-1mm}
\end{equation}
Besides the rescaled target $\mathbf v_m$, the read returns every other value,
weighted by its feature similarity $\langle\phi_n,\phi_m\rangle:=\phi_n^{\top}\phi_m$. 
These similarities cannot all vanish: at most $h$ features are mutually orthogonal in $\mathbb R^h$, and in practice the features share a large common component, so even unrelated tokens are similar (\cref{fig: relative_weihts}). 
As interference accumulates over all tokens, reads ``blur'' with more views and higher resolution. 
Memory also depends on chunk order: $\operatorname{Norm}$ rescales $\theta$ after each chunk, shrinking the influence of earlier chunks. 
As $\mathbf W_1,\mathbf W_3$ are updated too, each chunk is written with features shaped by the preceding ones. 
FSM's consolidation slows forgetting but fixes neither problem.

\subsection{Test-Time Solver}
\label{sec:tts}

Instead of performing a single optimization step per chunk, we propose a \emph{test-time solver} (TTS), which writes all inputs at once by solving the key--value fit exactly in closed form. 
This requires the map $f_\theta$ to be linear in the weights that are adapted. 
We therefore keep only $\mathbf W_2$ fast and make $\mathbf W_1,\mathbf W_3$ slow, which fixes the feature map $\phi$ of \cref{eq:swiglu} at inference, so the map becomes $f(\mathbf z)=\mathbf W_2\,\psi(\mathbf z)$ with $\psi(\mathbf z)=\phi(\mathbf z)/\|\phi(\mathbf z)\|$. 
Each input token thus carries a feature $\psi_n:=\psi(\hat{\mathbf k}_n)$, a value
$\mathbf v_n$, and a learnable weight $a_n>0$, as in FSM. 
Without the normalization, tokens with large $\|\phi_n\|$ would dominate the fit. 
With it, $a_n$ alone decides how much a token counts. 
We further center and rescale, $\tilde\psi_n=\mathbf S^{-1}(\psi_n-\bar\psi)$ and
$\tilde{\mathbf v}_n=\mathbf v_n-\bar{\mathbf v}$, where $\bar\psi$ and
$\bar{\mathbf v}$ are the $a_n$-weighted means and $\mathbf S$ is diagonal
with $S_{ii}^2=\sum_na_n(\psi_{n,i}-\bar\psi_i)^2$. Centering removes the mean that all features share, and rescaling gives every feature unit-weighted energy. The write then minimizes a weighted squared error with a penalty toward a learnable prior $\mathbf W_2^{\mathrm{init}}$,
\vspace{-2mm}
\begin{equation}
  \mathcal L_{\mathrm{TTS}}=\sum_na_n\big\|\mathbf W_2\tilde\psi_n-\tilde{\mathbf v}_n\big\|^2
  +\lambda\big\|\mathbf W_2-\mathbf W_2^{\mathrm{init}}\big\|_F^2 ,
  \label{eq:objective}
\vspace{-2mm}
\end{equation}
with a learnable strength $\lambda\ge10^{-3}$ that, after rescaling, acts
equally on all features. Unlike $\mathcal L_{\mathrm{FSM}}$,
this objective has a unique minimizer. The prior, learned across many scenes,
fills the feature directions that the inputs leave undetermined, which
matters when views are sparse. Every layer has its own TTS, with its own
$\psi$, $\mathbf W_2^{\mathrm{init}}$, and $\lambda$.

\vspace{0.5mm}
\noindent\textbf{Solve.}
Setting the gradient of \cref{eq:objective} to zero gives the minimizer
$\mathbf W_2^\star=\big(\tilde{\mathbf C}+\lambda\mathbf W_2^{\mathrm{init}}\big)\mathbf P^{-1}$
with $\mathbf P=\tilde{\mathbf G}+\lambda\mathbf I$, where
$\tilde{\mathbf G}=\sum_na_n\tilde\psi_n\tilde\psi_n^{\top}\in\mathbb R^{h\times h}$
and
$\tilde{\mathbf C}=\sum_na_n\tilde{\mathbf v}_n\tilde\psi_n^{\top}\in\mathbb R^{d\times h}$.
Since $\mathbf P\succeq\lambda\mathbf I$, it exists even for a single input
token. 

\vspace{0.5mm}
\noindent\textbf{Read.}
A token $\mathbf x_m$ reads the memory as
$\mathbf o_m=\bar{\mathbf v}+\mathbf W_2^\star\tilde\psi_m$, where its query
feature $\tilde\psi_m=\mathbf S^{-1}\big(\psi(\hat{\mathbf q}_m)-\bar\psi\big)$
is centered and rescaled like those of the input tokens.

\vspace{0.5mm}
\noindent\textbf{Interference suppression.}
Centering removes $\bar\psi$, the component shared by all keys that makes
even unrelated features similar, so this component now only adds the mean
value $\bar{\mathbf v}$ to every read. The solve suppresses the remaining
interference. Substituting $\mathbf W_2^\star$, a written key
$\hat{\mathbf k}_m$ reads
$\mathbf o_m=\bar{\mathbf v}
 +\textstyle\sum_na_n\langle\tilde\psi_n,\tilde\psi_m\rangle_{\mathbf P^{-1}}\tilde{\mathbf v}_n
 +\lambda\mathbf W_2^{\mathrm{init}}\mathbf P^{-1}\tilde\psi_m$, with
$\langle\tilde\psi_n,\tilde\psi_m\rangle_{\mathbf P^{-1}}:=\tilde\psi_n^{\top}\mathbf P^{-1}\tilde\psi_m$.
This is \cref{eq:readback} with $\langle\phi_n,\phi_m\rangle$ replaced by
the inner product in the metric $\mathbf P^{-1}$, the equivalent kernel of
ridge regression~\cite{bishop2006prml}. As $\tilde{\mathbf G}$ is the
weighted correlation matrix of the features, it is large along directions in
which many keys vary together.
$\mathbf P^{-1}$ down-weights these shared
directions and up-weights rare ones, which tell individual tokens apart. A
read thus returns mainly its own value and those of genuinely similar tokens
(\cref{fig: relative_weihts}). \Cref{sec:s_read} gives the full derivation of this
read.

\vspace{0.5mm}
\noindent\textbf{State.}
All quantities of the solve ($\bar\psi$, $\bar{\mathbf v}$, $\mathbf S$,
$\tilde{\mathbf G}$, $\tilde{\mathbf C}$) are closed-form functions of five
\emph{sufficient statistics}~\cite{fisher1922mathematical} of the raw pairs:
$\sum_na_n$, $\sum_na_n\psi_n$, $\sum_na_n\mathbf v_n$,
$\sum_na_n\psi_n\psi_n^{\top}$, and $\sum_na_n\mathbf v_n\psi_n^{\top}$, of
sizes $1$, $h$, $d$, $h\times h$, and $d\times h$ for any number of tokens
(see \cref{sec:s_sums}). As weighted sums, they can be accumulated chunk by chunk in any
order with the same result, so later frames never overwrite earlier ones
and, unlike in FSM, no consolidation is needed to prevent forgetting.

\vspace{0.5mm}
\noindent\textbf{Gradients.}
As $\mathbf P$ is symmetric positive definite, we factor
$\mathbf P=\mathbf L\mathbf L^{\top}$ by Cholesky and form $\mathbf L^{-1}$
once, so reads are matrix products, up to $10\times$ faster on GPUs than
triangular solves. The backward pass is also closed-form as the gradients of
$\tilde{\mathbf C}$, $\tilde{\mathbf G}$, $\mathbf W_2^{\mathrm{init}}$, and
$\lambda$ are matrix products, propagated to $\psi_n$, $\mathbf v_n$, and
$a_n$ through the sums. Unlike autograd, it stores only $\mathbf L^{-1}$ and
$\mathbf W_2^\star$ from the solve (see \cref{sec:s_backward}).
\begin{table*}[!thb]
\centering
\caption{\textbf{Quantitative ablations} on left-eye
Stereo4D~\cite{jin2025stereo4d}. We vary the number of input views
$V_{\mathrm{in}}$, shuffle the input order at $V_{\mathrm{in}}=32$ and raise
the resolution to $256\times256$; all other evaluations use $128\times128$.
\textbf{Ours} adds uncertainty maps (Unc.) to the TTS+RoPE variant, providing
per-pixel uncertainty at a cost of only 0.2\,dB at $V_{\mathrm{in}}=32$. Best
in bold, second-best underlined; see \cref{tab:s_arms} for complete results.}
\label{tab:pil_arms}
\vspace{-3.0mm}
\setlength{\tabcolsep}{3.6pt}
\resizebox{\linewidth}{!}{
\begin{tabular}{l*{12}{c}}
\toprule
\multirow{2}{*}{Methods} & $V_{\mathrm{in}}=4$ & $V_{\mathrm{in}}=8$ & $V_{\mathrm{in}}=16$ & \multicolumn{3}{c}{$V_{\mathrm{in}}=32$} & $V_{\mathrm{in}}=64$ & $V_{\mathrm{in}}=128$ & Shuffled & $256\times256$ & Train speed & Peak Mem. \\
\cmidrule(lr){2-2}\cmidrule(lr){3-3}\cmidrule(lr){4-4}\cmidrule(lr){5-7}\cmidrule(lr){8-8}\cmidrule(lr){9-9}\cmidrule(lr){10-10}\cmidrule(lr){11-11}\cmidrule(lr){12-12}\cmidrule(lr){13-13}
 & PSNR$\uparrow$ & PSNR$\uparrow$ & PSNR$\uparrow$ & PSNR$\uparrow$ & SSIM$\uparrow$ & LPIPS$\downarrow$ & PSNR$\uparrow$ & PSNR$\uparrow$ & PSNR$\uparrow$ & PSNR$\uparrow$ & s/step$\downarrow$ & GiB$\downarrow$ \\
\midrule
\textsc{FSM}-style~\cite{fsm}
& 22.19 & 25.96 & 30.43 & 33.63 & 0.948 & 0.029 & 29.50 & 21.89 & 30.94 & 18.84 & 22.34 & 66.81 \\
\textsc{MesaNet}-style~\cite{mesanet}
& 22.88 & 27.15 & 30.12 & 31.78 & 0.925 & 0.039 & 32.33 & 32.19 & 31.78 & 21.30 & 14.26 & 50.40 \\
\midrule
TTS+RoPE 
& \textbf{24.11} & 28.39 & \textbf{31.97} & \textbf{34.30} & \textbf{0.953} & \textbf{0.027} & \textbf{35.12} & \textbf{34.75} & \textbf{34.30} & 22.86 & 11.62 & 49.97 \\
\quad w/o. RoPE
& 23.56 & 28.26 & 31.65 & 33.81 & 0.951 & \underline{0.028} & 34.47 & \underline{34.39} & 33.81 & \textbf{23.42} & \textbf{11.32} & 49.97 \\
\quad w/o. time in RoPE
& 23.92 & \underline{28.46} & 31.74 & 33.87 & 0.949 & 0.029 & 34.70 & 33.83 & 33.87 & 22.63 & 11.62 & 49.97 \\
\quad w/o. prior $\mathbf W_2^{\mathrm{init}}$
& 23.04 & 27.21 & 31.28 & 33.67 & 0.947 & 0.030 & 34.18 & 34.14 & 33.67 & 21.70 & 11.61 & \textbf{48.28} \\
\quad w/o. learned $a_n$
& 23.33 & 28.22 & 31.78 & 34.06 & 0.951 & \underline{0.028} & 34.79 & 33.43 & 34.06 & 22.57 & \underline{11.36} & \underline{48.81} \\
\midrule
\rowcolor{LightCyan}
\textbf{Ours} (TTS+RoPE+Unc.)
& \underline{23.93} & \textbf{28.50} & \underline{31.90} & \underline{34.11} & \underline{0.952} & \underline{0.028} & \underline{34.95} & \underline{34.39} & \underline{34.11} & \underline{23.26} & 11.96 & 50.43 \\
\bottomrule
\end{tabular}
}
\vspace{-5mm}
\end{table*}

\subsection{Network Architecture}

\label{sec:arch}
Each view (\cref{fig: pipeline}), whether input or target, is converted into patch tokens. All
tokens pass through $L$ identical Transformer layers, and the query tokens
are decoded into pixels.

\vspace{0.5mm}
\noindent\textbf{Tokenization.} As in \textsc{FSM}, we compute per-pixel ray origins $\mathbf r^{\mathrm o}_i$ and
directions $\mathbf r^{\mathrm d}_i$ from $(\mathbf K_i,\mathbf E_i)$ and form
the ray map
$\mathrm{Concat}[\mathbf r^{\mathrm o}_i,\mathbf r^{\mathrm d}_i,\mathbf r^{\mathrm o}_i\times\mathbf r^{\mathrm d}_i]\in\mathbb R^{H\times W\times9}$. 
This gives pixel-aligned camera information for any intrinsics and pose. 
We stack it per pixel with the image $\mathcal I_i$ (scaled to $[-1,1]$;
zeros for target views) and a time map $\mathbf T_i\in\mathbb R^{H\times W\times1}$
holding $t_i$, min-max normalized to $[-1,1]$ over the input views (target
times alike). The resulting 13-channel map ($3+9+1$) is cut into non-overlapping $p\times p$ patches, each linearly projected from $13p^2$
values to a token $\mathbf x_n\in\mathbb R^d$. 
A view thus gives $HW/p^2$ tokens: with $p=8$, 32 views at $256\times256$ give 32{,}768 tokens and 64 views at $512\times512$ give 262{,}144, which makes full attention costly.

\vspace{0.5mm}
\noindent\textbf{Layers.}
Each of our $L$ layers applies three pre-norm residual sublayers in order:
\vspace{-2mm}
\begin{equation}
\begin{aligned}
  \mathbf X&\leftarrow\mathbf X+\operatorname{WinAttn}\big(\operatorname{LN}(\mathbf X)\big),\\
  \mathbf X&\leftarrow\mathbf X+\tts\big(\operatorname{LN}(\mathbf X)\big),\\
  \mathbf X&\leftarrow\mathbf X+\operatorname{MLP}\big(\operatorname{LN}(\mathbf X)\big),
\end{aligned}
\label{eq:block}
\vspace{-2mm}
\end{equation}
where $\mathbf X$ stacks tokens of all views, $\operatorname{LN}$ is layer
normalization~\cite{ba2016layer}, $\operatorname{WinAttn}$ is self-attention
within each view~\cite{zhang2026test,wang2026tttlrm}, and $\operatorname{MLP}$
is a per-token feed-forward network. 
Views only exchange information through the
\tts{} sublayers, and all other sublayers act within a single view. Finally, a linear layer with a sigmoid~\cite{rumelhart1986learning}
decodes each query token into the colors of its $p\times p$ patch, and the
patches of all tokens of a target view together form the rendered view
$\mathcal I_j$.

\vspace{0.5mm}
\noindent\textbf{\tts{} sublayer.}
Each token $\mathbf x_n$ is projected to $\mathbf q_n$, $\mathbf k_n$, $\mathbf v_n$ (\cref{sec:ttt}), and the learnable weight in \cref{eq:objective}, $a_n=\operatorname{softplus}(\mathbf w_a^{\top}\mathbf x_n+b_a)$, where $\mathbf w_a\in\mathbb R^d$ and $b_a\in\mathbb R$ are learned and initialized to give ${a_n\approx1}$. Input tokens add their feature $\psi_n$ of the rotated unit key (\cref{sec:rope}), value $\mathbf v_n$ and weight $a_n$ to the five weighted sums of \cref{sec:tts}. After one solve over all input tokens, every token $\mathbf x_m$, input or query, reads $\mathbf o_m$, which is RMS-normalized and projected by $\mathbf W_o$ as in \textsc{FSM}. Query tokens never write, so they affect neither the memory nor one another's outputs, and any number of target views can be rendered from one solve.

\vspace{0.5mm}
\noindent\textbf{Rendering loss.}
Each rendered target view $\mathcal I_j$ is compared with its ground truth
$\mathcal I_j^*$ using the loss of \textsc{FSM}~\cite{fsm}:
\vspace{-1mm}
\begin{equation}
  \mathcal L_{\mathrm{ren}}=\operatorname{MSE}(\mathcal I_j,\mathcal I_j^*)
  +0.5\,\operatorname{LPIPS}(\mathcal I_j,\mathcal I_j^*).
  \label{eq:ren}
\end{equation}

\subsection{2D-Time RoPE in TTS}
\label{sec:rope}

In a video, the inputs most relevant to a target patch are usually near it in
image position and time, but \tts{} relates tokens only through features in
which position and time are mixed with content. Rotary
embeddings~\cite{su2024roformer} give attention relative positions by
rotating the query and key before their inner product, along row, column, and
time in video generators~\cite{yang2025cogvideox,kong2024hunyuanvideo}. As
\tts{} has no such product, we rotate the unit query and key before its
feature map instead, using $\psi_n=\psi(\mathbf R_n\hat{\mathbf k}_n)$ for
input tokens and $\psi(\mathbf R_m\hat{\mathbf q}_m)$ for query tokens, with
$\mathbf R_n=\operatorname{diag}\big(\mathbf R(\bar r_n),\mathbf R(\bar c_n),\mathbf R(t_n)\big)$
for patch row $\bar r_n$, column $\bar c_n$ and time $t_n$, and
$\mathbf R(\cdot)$ the rotary matrix of \textsc{RoFormer}. Since
$\mathbf R_m^{\top}\mathbf R_n$ depends only on the offset between two
tokens, the same content at distant positions or times is rotated apart and
becomes distinguishable. The fit thus keeps a moving object's values at
different times apart, and a target patch favors nearby inputs, which keeps
neighboring patches coherent. Row and column are normalized by the grid size,
so every resolution spans the same angles, and $t$ is the normalized time of
\cref{sec:arch}.

\begin{figure*}[!t]
\centering
\vspace{-2mm}
\includegraphics[width=0.96\linewidth]{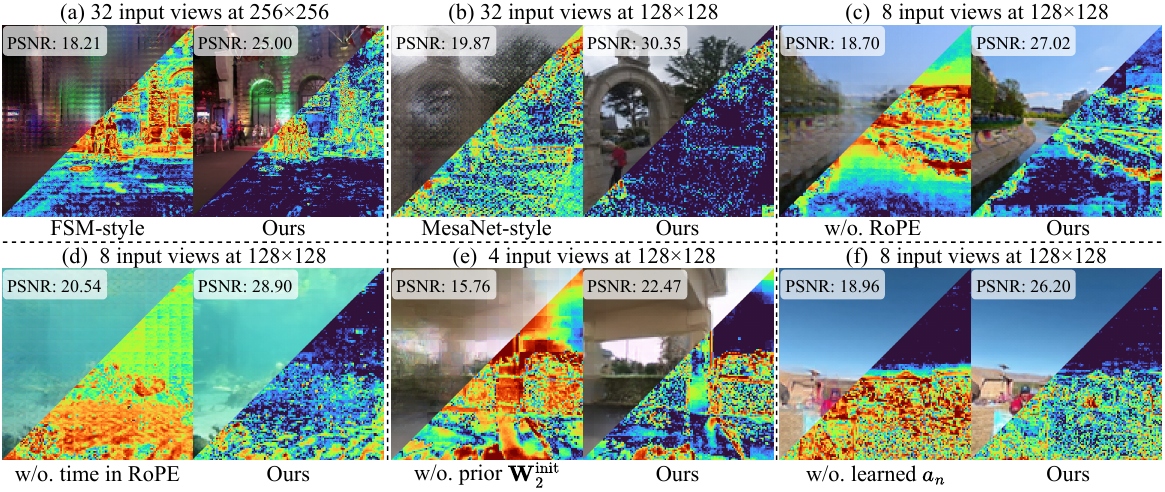}
\vspace{-4mm}
\caption{\textbf{Qualitative ablations} on left-eye Stereo4D. Each panel
compares one variant with our full model on one target view. In each image,
the upper-left half shows the rendering and the lower-right half its
absolute error (log scale, blue: low, red: high).}
\label{fig:ablations}
\vspace{-6mm}
\end{figure*}

\subsection{Uncertainty Maps}
\label{sec:confidence}

NVS renders images from the input views, so parts of the scene they do not depict must be inferred and may be wrong.  
Following \citet{kendall2017uncertainties}, we predict an uncertainty map that
combines an epistemic term, for what the inputs leave undetermined, and an
aleatoric term, for what remains hard to render even when they determine it, \eg, fine texture or reflections. 

\vspace{0.5mm}
\noindent\textbf{Epistemic.}
The inputs written to \tts{} also tell how reliable each read is. For any
noise scale $\sigma^2$, minimizing \cref{eq:objective} gives the maximum a
posteriori (MAP) estimate of $\mathbf W_2$ in the linear regression
$\tilde{\mathbf v}_n=\mathbf W_2\tilde\psi_n+\boldsymbol\varepsilon_n$ with
noise $\boldsymbol\varepsilon_n\sim\mathcal N(\mathbf 0,\sigma^2\mathbf I/a_n)$
and an isotropic prior with mean $\mathbf W_2^{\mathrm{init}}$ and variance
$\sigma^2/\lambda$~\cite{bishop2006prml}. Each row of $\mathbf W_2$ then has
posterior covariance $\sigma^2\mathbf P^{-1}$, and as the read
$\mathbf o_m=\bar{\mathbf v}+\mathbf W_2\tilde\psi_m$ is linear in
$\mathbf W_2$, it has covariance $\sigma^2u_m\mathbf I$ with
$u_m=\tilde\psi_m^{\top}\mathbf P^{-1}\tilde\psi_m=\|\mathbf L^{-1}\tilde\psi_m\|^2$.
This variance is small where many inputs have features along
$\tilde\psi_m$, and at most $\|\tilde\psi_m\|^2/\lambda$, reached where
$\tilde\psi_m$ is orthogonal to all of them, as we expect for surfaces hidden
from all inputs, regions outside all input views, or times far from all
inputs. We fuse $u_m$ of all layers, with gradients stopped, into a
log-variance per query token,
$\mathcal U^{\mathrm{epi}}_m=\beta_0+\textstyle\sum_{l=1}^{L}\beta_l\log(u_{m,l}/\bar u_l)$,
where $\bar u_l$ is a running mean of $u_{m,l}$, and $\beta_0$ (absorbing
$\sigma^2$) and $\beta_l\ge0$ are the only $L+1$ learnable parameters (see
\cref{sec:s_unc}). Bilinear upsampling to pixels gives the epistemic map
$\mathcal U^{\mathrm{epi}}$.

\vspace{0.5mm}
\noindent\textbf{Aleatoric.} The epistemic variance reflects only the input evidence, not how hard a
pixel is to render once the inputs show it, \eg, with fine texture,
reflections, or motion. A small head therefore maps each final query token,
detached so that the head leaves the Transformer unchanged, to a log-variance
correction at each of its $p\times p$ pixels, giving the aleatoric map
$\mathcal U^{\mathrm{ale}}$. The uncertainty map is
$\mathcal U=\operatorname{sg}[\mathcal U^{\mathrm{epi}}]+\mathcal U^{\mathrm{ale}}$,
where the stop-gradient $\operatorname{sg}$ keeps $\mathcal U^{\mathrm{epi}}$ fixed,
so $\mathcal U^{\mathrm{ale}}$ learns only the error that $\mathcal U^{\mathrm{epi}}$
leaves unexplained. It can thus correct $\mathcal U^{\mathrm{epi}}$ in both
directions: downward where the inputs miss a region that the learned prior
still renders correctly, and upward where the inputs show a region that the
model cannot render accurately. 

\vspace{0.5mm}
\noindent\textbf{Training.}
We use the training objective 
$\mathcal L_{\mathrm{unc}}=\mathcal L_{\mathrm{NLL}}(\mathcal U)+\mathcal L_{\mathrm{NLL}}(\mathcal U^{\mathrm{epi}})$,
where $\mathcal L_{\mathrm{NLL}}$ is the Gaussian negative
log-likelihood of the per-pixel squared error under a log-variance map (see
\cref{sec:s_unc}). Owing to the stop-gradient, only the second term trains $\beta_0$ and
$\beta_l$. 
Our final training objective is
$\mathcal L_{\mathrm{tot}}=\mathcal L_{\mathrm{ren}}+0.01\,\mathcal L_{\mathrm{unc}}$,
with $\mathcal L_{\mathrm{ren}}$ from \cref{eq:ren}.

\section{Ablation Studies} 
\label{sec:pilot}
\subsection{Setup}
\label{sec:pil_setup}
Before large-scale training, we run controlled ablations at a  moderate scale,
following \textsc{FSM}~\cite{fsm} (details in \cref{sec:s_ablation}).

\vspace{0.5mm}
\noindent\textbf{Variants.}
We compare TTS with two other memories:
an \textsc{FSM}-style memory~\cite{fsm}, written by one TTT step per chunk with
$\operatorname{Muon}$, $\operatorname{Norm}$, and consolidation, and a
\textsc{MesaNet}-style solver~\cite{mesanet}, which solves the regression exactly but
without our centering or prior. We also ablate the 2D-time RoPE
(\cref{sec:rope}), the learnable prior $\mathbf W_2^{\mathrm{init}}$ (set to
$\mathbf 0$), the token weights $a_n$ (set to 1), and the uncertainty head
(\cref{sec:confidence}). 
All variants use \textsc{FSM}'s backbone: $L=24$
layers, patch size $p=8$, token dimension $d=768$, and hidden dimension $h=1536$.

\vspace{0.5mm}
\noindent\textbf{Training.}
All variants, including baselines, are trained from scratch on left-eye
Stereo4D~\cite{jin2025stereo4d} videos for 32K steps on 8 H20 GPUs (batch
size 16 per GPU) with AdamW~\cite{loshchilov2019decoupled} at a peak
learning rate of $2\times10^{-4}$, using 32 input and 32 target views at
$128\times128$. 
Camera centers (relative to the mean input pose) and times are normalized to $[-1,1]$ over the inputs, so target views may fall outside this range. 
TTS uses PyTorch~\cite{paszke2019pytorch} matrix routines, with Triton~\cite{tillet2019triton} only for element-wise operations such as RoPE.

\vspace{0.5mm}
\noindent\textbf{Evaluation.}
From each of the 5{,}177 Stereo4D left-eye test clips with more than 136
frames, we sample 32 input and 8 target views at $128\times128$ and report
PSNR~\cite{chan1983hardware}, SSIM~\cite{wang2004image}, and
LPIPS~\cite{zhang2018unreasonable}. To test robustness without retraining,
we (i)~vary $V_{\mathrm{in}}$ from 4 (sparse) to 128 (dense), (ii)~shuffle
the order of the 32 input views, and (iii)~increase the resolution from
${128\times128}$, used in (i) and (ii), to $256\times256$, giving $4\times$
as many tokens per view. We also report training speed (wall-clock time per
step) and peak memory per GPU. We evaluate the uncertainty map by its Pearson
$L_1$~\cite{10.1098/rspl.1895.0041,galappaththige2026predictive}, the
pixel-wise correlation between the predicted standard deviation
$\exp(\mathcal U/2)$ and the absolute rendering error.

\subsection{Analysis}
\noindent\textbf{TTS.}
When training on 32 views at $128\times128$, the \textsc{FSM}-style memory is
only slightly behind our TTS variants (\cref{tab:pil_arms}). 
As it writes in fixed-size chunks, more tokens mean more updates and more accumulated
interference, so the gap widens with more views and at a higher resolution
(\cref{fig:ablations}~(a)). It also loses 2.7\,dB PSNR when the inputs are
shuffled, as its memory depends on chunk order, whereas both closed-form
solvers write all inputs at once and are unaffected. The \textsc{MesaNet}-style solver
stays over 2\,dB below ours from 32 views on and renders visibly blurrier
(\cref{fig:ablations}~(b)), mainly because, without centering, the component
shared by all keys blends all values into every read. 
Our solver also halves the time per training step of the \textsc{FSM}-style memory and cuts its peak memory
by a quarter. 
The latter updates its chunks sequentially, back-propagating
through Newton--Schulz iterations, whereas we accumulate the sums in
parallel, and both the solve and its gradient are a few dense matrix
products, which GPUs run efficiently.

\begin{table}[!t]
\centering
\caption{\textbf{Quantitative uncertainty ablations} on left-eye Stereo4D~\cite{jin2025stereo4d}. The bottom row correlates the epistemic and aleatoric maps with each other, not with the absolute error.}
\label{tab:uncertainty}
\vspace{-3.5mm}
\resizebox{\linewidth}{!}{
\begin{tabular}{l|cccccc}
\toprule
\multirow{2}{*}{Maps} & \multicolumn{6}{c}{$V_{\mathrm{in}}$ at $128\times128$ (Pearson $L_1\uparrow$)}\\
\cline{2-7}
& 4 & 8 & 16 & 32 & 64 & 128 \\
\midrule
Epistemic $\mathcal U^{\mathrm{epi}}$ & 0.394 & 0.414 & 0.390 & 0.364 & 0.359 & 0.386 \\
Aleatoric $\mathcal U^{\mathrm{ale}}$  & 0.253 & 0.342 & 0.406 & 0.434 & 0.441 & 0.459 \\
\rowcolor{LightCyan}
\textbf{Uncertainty $\mathcal U$} & \textbf{0.430} & \textbf{0.531} & \textbf{0.575} & \textbf{0.594} & \textbf{0.602} & \textbf{0.603} \\
\midrule
$\mathcal U^{\mathrm{epi}}$ and $\mathcal U^{\mathrm{ale}}$ & 0.006 & 0.059 & 0.057 & 0.022 & $-$0.005 & 0.067 \\
\bottomrule
\end{tabular}
}
\vspace{-6mm}
\end{table}

\begin{table}[!b]
\centering
\vspace{-5mm}
\caption{\textbf{Quantitative 4D NVS comparisons.} Our method is best on all datasets, each at a different resolution, and for every
$V_{\mathrm{in}}$.}
\label{tab:4d}
\vspace{-3mm}
\setlength{\tabcolsep}{2.5pt}
\resizebox{\linewidth}{!}{
\begin{tabular}{@{}cl*{9}{c}@{}}
\toprule
& & \multicolumn{3}{c}{\textbf{Stereo4D}} & \multicolumn{3}{c}{\textbf{DyCheck} (OOD)} & \multicolumn{3}{c}{\textbf{NVIDIA} (OOD)} \\
\cmidrule(lr){3-5}\cmidrule(lr){6-8}\cmidrule(l){9-11}
\multirow{-2}{*}{$V_{\mathrm{in}}$} & \multirow{-2}{*}{Methods} & {\scriptsize PSNR$\uparrow$} & {\scriptsize SSIM$\uparrow$} & {\scriptsize LPIPS$\downarrow$} & {\scriptsize PSNR$\uparrow$} & {\scriptsize SSIM$\uparrow$} & {\scriptsize LPIPS$\downarrow$} & {\scriptsize PSNR$\uparrow$} & {\scriptsize SSIM$\uparrow$} & {\scriptsize LPIPS$\downarrow$} \\
\midrule
 & \textsc{4DGT}~\cite{4dgt} & 16.27 & 0.567 & 0.418 & 9.43 & 0.301 & 0.561 & 14.64 & 0.360 & 0.696 \\
 & \textsc{MoVieS}~\cite{movies} & \underline{22.26} & \underline{0.706} & \underline{0.240} & 16.83 & 0.485 & 0.403 & 15.88 & 0.349 & \underline{0.403} \\
 & \textsc{NeoVerse}~\cite{neoverse} & 16.15 & 0.469 & 0.461 & 16.28 & 0.395 & 0.437 & 11.19 & 0.176 & 0.654 \\
 & \textsc{FSM}~\cite{fsm} & 19.49 & 0.595 & 0.388 & \underline{19.15} & \underline{0.511} & \underline{0.316} & \underline{18.23} & 0.354 & 0.459 \\
 & \textsc{C4G}~\cite{c4g} & 17.36 & 0.538 & 0.562 & 16.05 & 0.439 & 0.638 & 15.23 & \underline{0.373} & 0.895 \\
 & \textsc{FL4N}~\cite{fl4n} & 21.43 & 0.656 & 0.300 & 17.34 & 0.447 & 0.372 & 17.58 & 0.356 & 0.479 \\
\rowcolor{LightCyan}
\cellcolor{white}\multirow{-7}{*}{16} & \textbf{Ours} & \textbf{25.88} & \textbf{0.805} & \textbf{0.156} & \textbf{20.55} & \textbf{0.602} & \textbf{0.247} & \textbf{22.89} & \textbf{0.665} & \textbf{0.155} \\
\midrule
 & \textsc{4DGT}~\cite{4dgt} & 20.39 & 0.641 & 0.352 & 13.73 & 0.431 & 0.469 & 14.32 & 0.363 & 0.734 \\
 & \textsc{MoVieS}~\cite{movies} & \underline{22.07} & \underline{0.703} & \underline{0.241} & 17.12 & 0.491 & 0.400 & 15.50 & 0.334 & \underline{0.432} \\
 & \textsc{NeoVerse}~\cite{neoverse} & 16.17 & 0.470 & 0.458 & 17.12 & 0.418 & 0.404 & 11.62 & 0.188 & 0.630 \\
 & \textsc{FSM}~\cite{fsm} & 20.44 & 0.615 & 0.356 & \underline{20.71} & \underline{0.571} & \underline{0.267} & \underline{18.18} & 0.334 & 0.487 \\
 & \textsc{C4G}~\cite{c4g} & 17.33 & 0.537 & 0.571 & 16.30 & 0.442 & 0.639 & 15.11 & \underline{0.374} & 0.913 \\
 & \textsc{FL4N}~\cite{fl4n} & 21.08 & 0.644 & 0.308 & 17.98 & 0.456 & 0.346 & 17.55 & 0.357 & 0.480 \\
\rowcolor{LightCyan}
\cellcolor{white}\multirow{-7}{*}{32} & \textbf{Ours} & \textbf{26.02} & \textbf{0.808} & \textbf{0.153} & \textbf{22.15} & \textbf{0.643} & \textbf{0.207} & \textbf{23.50} & \textbf{0.676} & \textbf{0.147} \\
\midrule
 & \textsc{4DGT}~\cite{4dgt} & 20.98 & 0.651 & 0.346 & 18.62 & \underline{0.530} & 0.400 & 13.98 & 0.364 & 0.764 \\
 & \textsc{MoVieS}~\cite{movies} & \underline{21.87} & \underline{0.700} & \underline{0.244} & 17.07 & 0.496 & 0.400 & 15.49 & 0.336 & \underline{0.437} \\
 & \textsc{NeoVerse}~\cite{neoverse} & 16.20 & 0.470 & 0.456 & 17.74 & 0.433 & 0.375 & 11.66 & 0.192 & 0.622 \\
 & \textsc{FSM}~\cite{fsm} & 19.36 & 0.560 & 0.401 & \underline{19.17} & 0.467 & 0.364 & 17.05 & 0.282 & 0.629 \\
 & \textsc{C4G}~\cite{c4g} & 17.17 & 0.534 & 0.582 & 16.31 & 0.446 & 0.646 & 14.62 & \underline{0.370} & 0.929 \\
 & \textsc{FL4N}~\cite{fl4n} & 20.81 & 0.635 & 0.314 & 18.81 & 0.481 & \underline{0.315} & \underline{17.49} & 0.359 & 0.489 \\
\rowcolor{LightCyan}
\cellcolor{white}\multirow{-7}{*}{64} & \textbf{Ours} & \textbf{26.06} & \textbf{0.809} & \textbf{0.153} & \textbf{23.72} & \textbf{0.683} & \textbf{0.173} & \textbf{23.76} & \textbf{0.684} & \textbf{0.145} \\
\bottomrule
\end{tabular}
}
\end{table}

\vspace{0.5mm}
\noindent\textbf{RoPE.}
2D-time RoPE raises PSNR at 32 views from 33.81 to 34.30\,dB, 0.43\,dB of it
from the time axis, and reduces grid artifacts by coupling neighboring
patches (\cref{fig:ablations}~(c,d)). At $256\times256$, however, it costs
0.56\,dB. 
Coordinates are normalized to the patch grid, so neighboring
patches are closer in angle than at the training resolution. Our large-scale
training therefore varies the resolution.

\vspace{0.5mm}
\noindent\textbf{Prior and token weights.}
The learnable prior $\mathbf W_2^{\mathrm{init}}$ helps when inputs are sparse. 
Without it, blocky artifacts appear with 4 inputs (\cref{fig:ablations}~(e)).
The learnable token weights $a_n$ let the solve weight tokens differently, which
recovers more detail, especially in textured regions
(\cref{fig:ablations}~(f)).

\begin{figure}[!t]
\centering
\includegraphics[width=\linewidth]{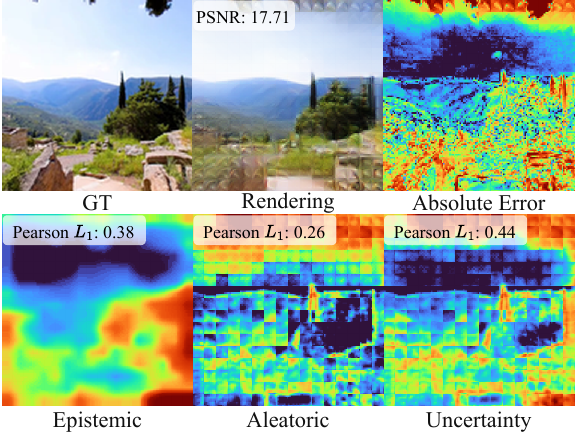}
\vspace{-7.5mm}
\caption{\textbf{Qualitative uncertainty ablations} on left-eye Stereo4D, rendered from 4 input views at
$128\times128$. The aleatoric and total uncertainty maps show a patch grid because each token is decoded
independently into an $8\times8$ patch, whereas the epistemic map is
bilinearly upsampled.}
\label{fig:ablate_uncertainty}
\vspace{-6mm}
\end{figure}

\vspace{0.5mm}
\noindent\textbf{Uncertainty.}
Adding the uncertainty map costs only 0.2\,dB at 32 views, as the head
reads detached tokens. \Cref{tab:uncertainty} evaluates each map by
Pearson $L_1$. With few inputs, the epistemic map explains the error better
(\cref{fig:ablate_uncertainty}), as most errors lie where the inputs give
little evidence. From 16 views on, the aleatoric map takes over. 
Once the inputs cover the scene, the remaining errors come from content that is hard
to render, such as fine texture. The two maps are nearly uncorrelated and
thus complementary, and their combination $\mathcal U$ explains the
error best at every number of views. 
We therefore keep them in large-scale
training.

\definecolor{LightCyan}{rgb}{0.88,1,1}

\definecolor{LightCyan}{rgb}{0.88,1,1}
\begin{figure*}[!htb]
\centering
\includegraphics[width=0.97\linewidth]{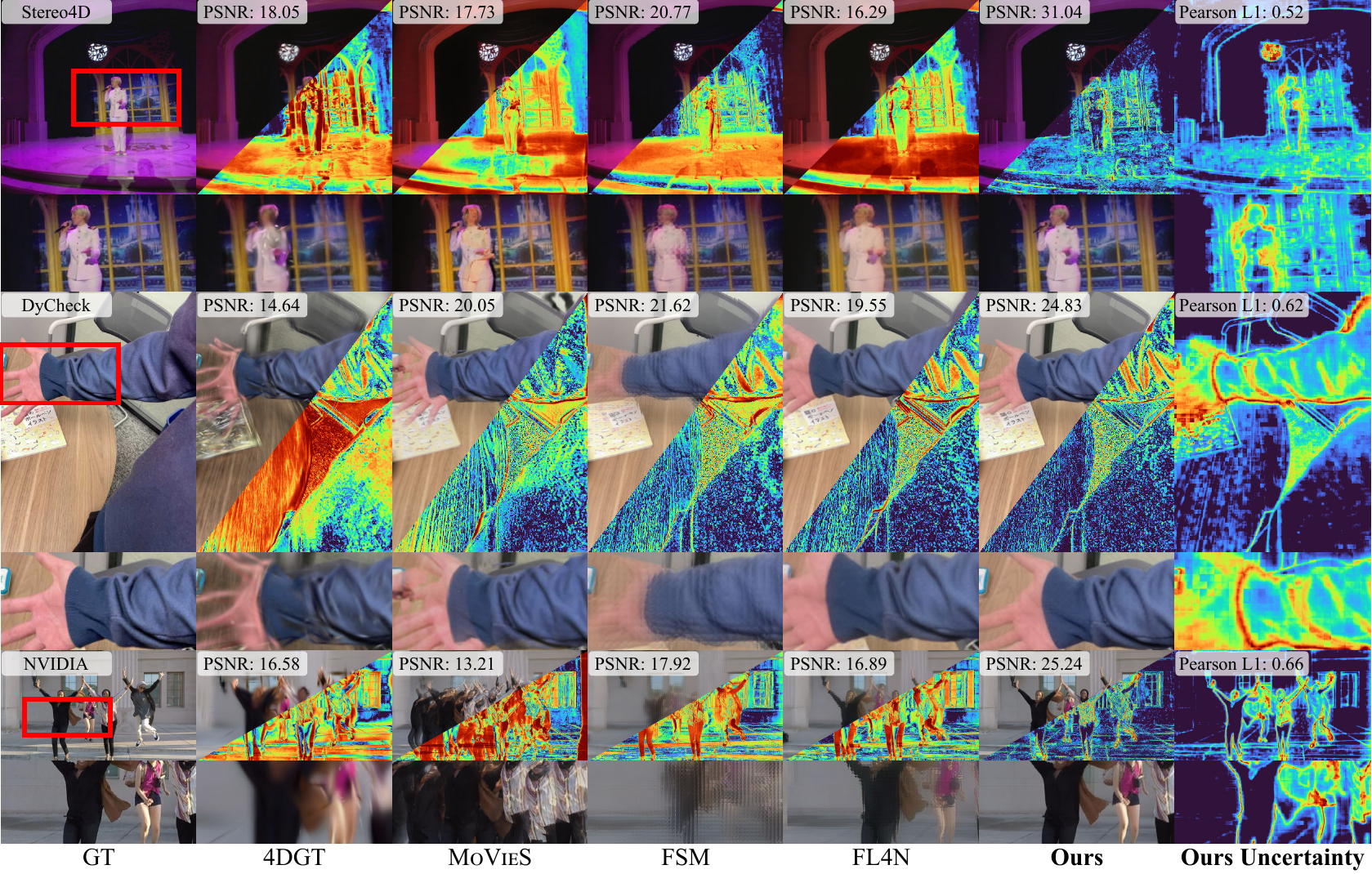}
\vspace{-4mm}
\caption{\textbf{Qualitative 4D NVS comparisons} with $V_{\mathrm{in}}=32$ against four representative baselines (\cref{tab:4d}): a \Singer{} under changing stage light (Stereo4D), a waving \Hand{} (DyCheck), and people \Jumping{} (NVIDIA). Top rows: renderings (upper left) and absolute errors (lower right). 
Bottom rows: zoom-ins of the red boxes. Last column: our uncertainty, rank-matched to the error. Blue: low, red: high.}
\label{fig:4d_qual}
\vspace{-5mm}
\end{figure*}

\vspace{-1mm}
\section{Large-Scale Experiments}
\vspace{-1mm}
\label{sec:exp}

\subsection{Setup}
\vspace{-0.5mm}
\label{sec:exp_setup}
\noindent\textbf{Training data.}
We scale our model from \cref{sec:pilot} to a mixture of 40 public data
sources (\cref{sec:s_data}) that spans monocular dynamic videos (\eg,
SpatialVID~\cite{wang2025spatialvid}, EgoDex~\cite{hoque2025egodex}),
multi-camera captures (\eg, Waymo~\cite{sun2020waymo},
Stereo4D~\cite{jin2025stereo4d}), static scenes (\eg,
DL3DV~\cite{ling2024dl3dv}, RealEstate10K~\cite{zhou2018stereo}) and
synthetic renders (\eg, Kubric~\cite{greff2022kubric},
Hypersim~\cite{roberts2021hypersim}). Every 1K steps, we reweight the sources
by their training PSNR.

\vspace{0.5mm}
\noindent\textbf{Sampling.}
Training views are sampled with a curriculum. Each step draws 4--256 input and
6--64 target views from a temporal window of 2.5--24$\times$ the number of inputs
(capped at the whole sequence). The number of views, the window, and the input
resolution, which starts from 128 pixels, all grow gradually during
training, and the batch size adapts to a fixed token budget per GPU. Input cameras
and timestamps are perturbed with noise (see \cref{sec:s_sampling} for details).

\vspace{0.5mm}
\noindent\textbf{Optimization.}
We use the same architecture and optimizer as in the ablations and evaluate an
exponential moving average (EMA)~\cite{polyak1992acceleration} of the weights.
Training takes 18 days on 32 H20 GPUs, with only 36\% of \textsc{FSM}'s training
tokens.

\vspace{-1.5mm}
\subsection{4D NVS Performance}
\label{sec:exp_protocol}
\vspace{-2mm}
\noindent\textbf{Test sets.}
On the Stereo4D test clips, we input left-eye views at the native
$512\times512$ and score the right eye, which is harder than scoring the input
eye as in the ablations. On two out-of-domain (OOD) datasets, we score the
frames of DyCheck~\cite{gao2022dynamic} ($360\times480$, 14 phone sequences)
between inputs sampled along the trajectory, and camera~1 of
NVIDIA~\cite{yoon2020novel} ($960\times540$, 9 sequences from a 12-camera rig)
at all 12 instants as in \mbox{\textsc{DynamicNeRF}}~\cite{gao2021dynamic}, with inputs
from all instants of the other 11 cameras. We use
$V_{\mathrm{in}}\in\{16,32,64\}$. \Cref{sec:s_4d} adds bullet-time, resolution,
novel-view and time-interpolation tests.

\vspace{0.0mm}
\noindent\textbf{Baselines.} 
We compare with the six recent feed-forward 4D NVS methods:
\textsc{4DGT}~\cite{4dgt}, \textsc{MoVieS}~\cite{movies}, \textsc{NeoVerse}~\cite{neoverse},
\textsc{C4G}~\cite{c4g}, \textsc{FL4N}~\cite{fl4n}, and \textsc{FSM}~\cite{fsm}, each run from its best
released checkpoint on the same inputs. As \textsc{C4G} takes no cameras, we apply its
test-time pose alignment~\cite{c4g}. \Cref{sec:s_baselines} compares their training data,
resolutions, model sizes, and runtimes with ours.

\vspace{0.0mm}
\noindent\textbf{Analysis.}
Our model performs best (\cref{tab:4d,fig:4d_qual}) and keeps improving as
$V_{\mathrm{in}}$ increases, whereas \textsc{FSM} degrades from 32 to 64 views as the interference accumulates. The baselines fail for different reasons. \textsc{4DGT}
keeps each Gaussian near its input time, leaving holes at other times (\Hand). \textsc{MoVieS}
moves its Gaussians over time, but they fracture under large non-rigid motion
(\Jumping), and their color and opacity are fixed, so the light on the \Singer{} turns
red instead of purple. 
\textsc{FL4N} adapts a time-invariant residual to the target time only through a light
modulation, so it blurs the motion (\Jumping) and misses the lighting change (\Singer). Our implicit memory, solved by TTS
from all inputs, restores both. Our uncertainty also follows the error closely, peaking at the edges of moving objects
such as the outline of the \Hand{} (quantitative results in \cref{sec:s_unc4d}). Finally, without any retraining, our 4D model is also competitive with SOTA 3D NVS
models on static scenes (\cref{sec:s_3d}).
\vspace{-5mm}
\section{Conclusion}
\vspace{-2mm}
Transformer-based 4D NVS models struggle to scale to many input views at high
resolution. 
We address this with a test-time solver layer that writes all input tokens into a single fast weight in closed form. 
We show that accurately writing tokens into memory matters as much as designing a more complex memory. 
Our approach predicts a feed-forward uncertainty map that tracks per-pixel rendering errors, which can be used for human inspection and downstream machine perception. 
Finally, we believe our solver may also extend to other long-sequence tasks, such as video generation, language modeling, and robot manipulation.

{
    \small
    \bibliographystyle{ieeenat_fullname}
    \bibliography{main}
}

\clearpage
\maketitlesupplementary
\appendix
\setcounter{table}{0}
\renewcommand{\thetable}{A\arabic{table}}
\setcounter{figure}{0}
\renewcommand{\thefigure}{A\arabic{figure}}
\setcounter{equation}{0}
\renewcommand{\theequation}{A\arabic{equation}}

\setcounter{equation}{0}
\renewcommand{\theequation}{A\arabic{equation}}

\section*{Overview}
This supplementary material provides additional details of our method and
additional results; its sections can be read independently.
\Cref{sec:s_method} derives the read of our Test-Time Solver (TTS), its
fixed-size state of five weighted sums, and its closed-form backward pass
(\cref{sec:tts}), as well as the uncertainty maps (\cref{sec:confidence}).
\Cref{sec:s_ablation} gives the setup and the complete results of the
ablations (\cref{sec:pil_setup}). \Cref{sec:s_train} describes the training data, the
sampling during training, and the optimization of our large-scale model
(\cref{sec:exp_setup}). \Cref{sec:s_baselines} compares the training setups and runtimes
of the 4D baselines with ours, and \cref{sec:s_4d} provides further 4D
evaluations (\cref{sec:exp_protocol}). \Cref{sec:s_3d} reports static 3D NVS and
uncertainty, \cref{sec:s_active} applies our uncertainty to active view
selection, and \cref{sec:s_grid} analyzes the grid artifacts. We encourage
readers to watch the \textbf{accompanying video} for the dynamic results and
to explore our code repository (linked from the project page), which contains the implementation of our TTS
layers.

\section{Additional Method Details}
\label{sec:s_method}

\subsection{Derivation of the TTS Read}
\label{sec:s_read}
We derive the read step of \cref{sec:tts}. For reference, we first recall the
quantities of one TTS layer. Each input token $n=1,\dots,N$ writes its
feature $\psi_n=\psi(\mathbf R_n\hat{\mathbf k}_n)\in\mathbb R^h$, \ie, the
feature map of its rotated unit key, its value
$\mathbf v_n\in\mathbb R^d$, and its weight
$a_n>0$, all
predicted from the token $\mathbf x_n$ by learnable projections. Each reading
token $m$, input or query, reads with its feature
$\psi(\hat{\mathbf q}_m)\in\mathbb R^h$, short for
$\psi(\mathbf R_m\hat{\mathbf q}_m)$. The fast weight is
$\mathbf W_2\in\mathbb R^{d\times h}$. Besides these projections, the solve
has two learnable parameters: the prior
$\mathbf W_2^{\mathrm{init}}\in\mathbb R^{d\times h}$ and the ridge
strength $\lambda>0$. The objective of \cref{eq:objective},
\begin{equation}
\mathcal L_{\mathrm{TTS}}(\mathbf W_2)
=\textstyle\sum_na_n\|\mathbf W_2\tilde\psi_n-\tilde{\mathbf v}_n\|^2
+\lambda\|\mathbf W_2-\mathbf W_2^{\mathrm{init}}\|_F^2 ,
\label{eq:s_obj}
\end{equation}
is a weighted linear least-squares problem with a ridge term, which pulls
$\mathbf W_2$ toward the learnable prior instead of toward zero. It uses the
weighted means $\bar\psi\in\mathbb R^h$ and $\bar{\mathbf v}\in\mathbb R^d$,
\begin{equation}
\bar\psi=\frac{\sum_na_n\psi_n}{\sum_na_n},\qquad
\bar{\mathbf v}=\frac{\sum_na_n\mathbf v_n}{\sum_na_n},
\label{eq:s_means}
\end{equation}
the centered and rescaled features and values
\begin{equation}
\begin{aligned}
&\tilde\psi_n=\mathbf S^{-1}(\psi_n-\bar\psi),\qquad
\tilde{\mathbf v}_n=\mathbf v_n-\bar{\mathbf v},\\
&S_{ii}^2=\textstyle\sum_na_n(\psi_{n,i}-\bar\psi_i)^2,
\end{aligned}
\label{eq:s_center}
\end{equation}
with $\mathbf S\in\mathbb R^{h\times h}$ diagonal, and the matrices
$\tilde{\mathbf G},\mathbf P\in\mathbb R^{h\times h}$ and
$\tilde{\mathbf C}\in\mathbb R^{d\times h}$,
\begin{equation}
\tilde{\mathbf G}=\textstyle\sum_na_n\tilde\psi_n\tilde\psi_n^{\top},\quad
\tilde{\mathbf C}=\textstyle\sum_na_n\tilde{\mathbf v}_n\tilde\psi_n^{\top},\quad
\mathbf P=\tilde{\mathbf G}+\lambda\mathbf I .
\label{eq:s_GC}
\end{equation}

\noindent\textbf{Solve.} Expanding the squared norms of \cref{eq:s_obj} and
collecting the terms with \cref{eq:s_GC},
\begin{equation}
\mathcal L_{\mathrm{TTS}}
=\operatorname{tr}\big(\mathbf W_2\mathbf P\mathbf W_2^{\top}\big)
-2\operatorname{tr}\big(\mathbf W_2(\tilde{\mathbf C}+\lambda\mathbf W_2^{\mathrm{init}})^{\top}\big)
+\mathrm{const},
\label{eq:s_quad}
\end{equation}
where $\mathrm{const}$ does not depend on $\mathbf W_2$. The gradient is
$2\mathbf W_2\mathbf P-2(\tilde{\mathbf C}+\lambda\mathbf W_2^{\mathrm{init}})$.
Since $\mathbf P\succeq\lambda\mathbf I\succ0$, \cref{eq:s_quad} is strictly
convex, and its gradient vanishes at the unique minimizer
\begin{equation}
\mathbf W_2^{\star}=\big(\tilde{\mathbf C}+\lambda\mathbf W_2^{\mathrm{init}}\big)\mathbf P^{-1}.
\label{eq:s_solve}
\end{equation}

\noindent\textbf{Read.} A token reads
$\mathbf o_m=\bar{\mathbf v}+\mathbf W_2^{\star}\tilde\psi_m$. Inserting
\cref{eq:s_solve} and then the definition of $\tilde{\mathbf C}$
(\cref{eq:s_GC}),
\begin{equation*}
\begin{aligned}
\mathbf o_m
&=\bar{\mathbf v}+\big(\tilde{\mathbf C}+\lambda\mathbf W_2^{\mathrm{init}}\big)\mathbf P^{-1}\tilde\psi_m\\
&=\bar{\mathbf v}+\tilde{\mathbf C}\mathbf P^{-1}\tilde\psi_m
+\lambda\mathbf W_2^{\mathrm{init}}\mathbf P^{-1}\tilde\psi_m\\
&=\bar{\mathbf v}+\textstyle\sum_na_n\tilde{\mathbf v}_n\big(\tilde\psi_n^{\top}\mathbf P^{-1}\tilde\psi_m\big)
+\lambda\mathbf W_2^{\mathrm{init}}\mathbf P^{-1}\tilde\psi_m .
\end{aligned}
\end{equation*}
The factor $\tilde\psi_n^{\top}\mathbf P^{-1}\tilde\psi_m=:\langle\tilde\psi_n,\tilde\psi_m\rangle_{\mathbf P^{-1}}$
is a scalar, so it moves in front of $\tilde{\mathbf v}_n$, which gives the
read of \cref{sec:tts}:
\begin{equation}
\mathbf o_m=\bar{\mathbf v}
+\textstyle\sum_na_n\langle\tilde\psi_n,\tilde\psi_m\rangle_{\mathbf P^{-1}}\tilde{\mathbf v}_n
+\lambda\mathbf W_2^{\mathrm{init}}\mathbf P^{-1}\tilde\psi_m .
\label{eq:s_read}
\end{equation}
Its three terms are the mean value, the written values weighted by the
similarity of their features to $\tilde\psi_m$ under $\mathbf P^{-1}$, and
the contribution of the prior $\mathbf W_2^{\mathrm{init}}$.

\subsection{A Fixed-Size State for Any Number of Tokens}
\label{sec:s_sums}
We show that $\mathbf W_2^{\star}$ depends on the input tokens only through
the five sums $\sum_na_n$, $\sum_na_n\psi_n$, $\sum_na_n\mathbf v_n$,
$\sum_na_n\psi_n\psi_n^{\top}$, and $\sum_na_n\mathbf v_n\psi_n^{\top}$.
Substituting \cref{eq:s_center} into \cref{eq:s_GC} gives
$\tilde{\mathbf G}=\mathbf S^{-1}\mathbf G'\mathbf S^{-1}$ and
$\tilde{\mathbf C}=\mathbf C'\mathbf S^{-1}$ with the centered sums
\begin{equation}
\begin{aligned}
\mathbf G'&:=\textstyle\sum_na_n(\psi_n-\bar\psi)(\psi_n-\bar\psi)^{\top}\\
&\phantom{:}=\textstyle\sum_na_n\psi_n\psi_n^{\top}-\bar\psi\bar\psi^{\top}\sum_na_n ,\\
\mathbf C'&:=\textstyle\sum_na_n(\mathbf v_n-\bar{\mathbf v})(\psi_n-\bar\psi)^{\top}\\
&\phantom{:}=\textstyle\sum_na_n\mathbf v_n\psi_n^{\top}-\bar{\mathbf v}\bar\psi^{\top}\sum_na_n ,
\end{aligned}
\label{eq:s_brackets}
\end{equation}
where the second forms follow by multiplying out and using
$\sum_na_n\psi_n=\bar\psi\sum_na_n$ and
$\sum_na_n\mathbf v_n=\bar{\mathbf v}\sum_na_n$ (\cref{eq:s_means}). By
\cref{eq:s_center}, the diagonal $\mathbf S$ is given by the diagonal of
$\mathbf G'$:
\begin{equation}
S_{ii}^2=(\mathbf G')_{ii}
=\textstyle\sum_na_n\psi_{n,i}^2-\bar\psi_i^2\sum_na_n .
\label{eq:s_S}
\end{equation}
With \cref{eq:s_solve}, the solution is therefore
\begin{equation}
\mathbf W_2^{\star}
=\big(\mathbf C'\mathbf S^{-1}+\lambda\mathbf W_2^{\mathrm{init}}\big)
\big(\mathbf S^{-1}\mathbf G'\mathbf S^{-1}+\lambda\mathbf I\big)^{-1}.
\label{eq:s_w_sums}
\end{equation}
Its right-hand side contains only the five sums, through $\mathbf G'$,
$\mathbf C'$, $\mathbf S$ and the means, and the learnable $\lambda$ and
$\mathbf W_2^{\mathrm{init}}$. The same holds for every read
$\mathbf o_m=\bar{\mathbf v}+\mathbf W_2^{\star}\tilde\psi_m$. The sums have
sizes $1$, $h$, $d$, $h\times h$, and $d\times h$ for any number of tokens,
and each adds one term per token, so chunks can be accumulated in any order
with the same result.

None of these quantities is a trained parameter. The five sums, the means,
$\mathbf S$, $\mathbf P$, its Cholesky factor $\mathbf L$, and
$\mathbf W_2^{\star}$ are recomputed from the tokens of every input
sequence, during training and at test time alike, so the solve is a
differentiable layer. Training updates only learnable parameters: the prior
$\mathbf W_2^{\mathrm{init}}$, the ridge strength $\lambda$, and the
projections that produce $\psi_n$, $\mathbf v_n$, $a_n$, and the reading features; at test time, all of them are
frozen.

\subsection{Closed-Form Backward Pass}
\label{sec:s_backward}
\begin{figure*}[!t]
\centering
\includegraphics[width=0.8\linewidth]{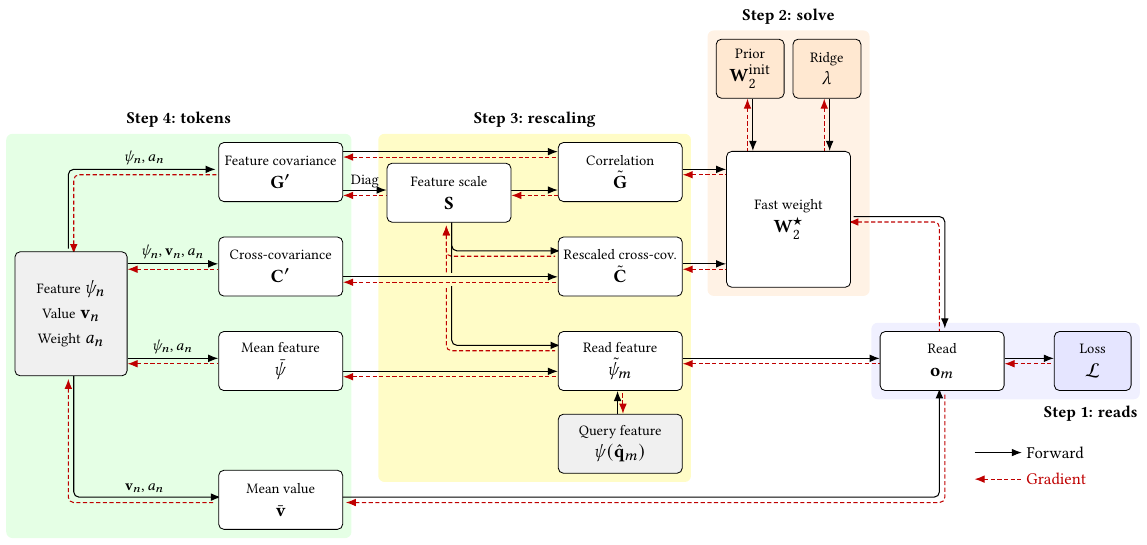}
\vspace{-2mm}
\caption{\textbf{Computation graph of one TTS layer.} Black arrows show the
forward pass from the input tokens (left) to the loss (right). Dashed red
arrows show the gradients, which the backward pass (\cref{sec:s_backward})
propagates in the reverse direction, one colored block per step. Gray:
inputs of the layer; orange: learnable parameters; $m$ indexes the tokens
that read.}
\label{fig:s_chain}
\vspace{-5mm}
\end{figure*}

Here we derive the backward pass of one TTS layer (paragraph ``Gradients'' of
\cref{sec:tts}). Given the gradients of the training loss $\mathcal L$ with
respect to the reads $\mathbf o_m$, it follows the dashed arrows of
\cref{fig:s_chain}, one colored block per step, and returns the
gradients of $\mathbf W_2^{\mathrm{init}}$, $\lambda$, $\psi_n$,
$\mathbf v_n$, and $a_n$.

\noindent\textbf{Notation.} We write
$\nabla_{\mathbf M}:=\partial\mathcal L/\partial\mathbf M$ and
$\langle\mathbf M,\mathbf Y\rangle:=\operatorname{tr}(\mathbf M^{\top}\mathbf Y)$.
If $\mathcal L$ depends on some quantities only through
$\mathbf M_1,\mathbf M_2,\dots$, then
$\delta\mathcal L=\sum_k\langle\nabla_{\mathbf M_k},\delta\mathbf M_k\rangle$.
In each step, we insert the differentials of the forward relations, move all
factors away from the differentials with
\begin{equation}
\langle\mathbf M,\mathbf Y\mathbf Q\rangle=\langle\mathbf M\mathbf Q^{\top},\mathbf Y\rangle,
\quad
\langle\mathbf M,\mathbf Q\mathbf Y\rangle=\langle\mathbf Q^{\top}\mathbf M,\mathbf Y\rangle ,
\label{eq:s_b_move}
\end{equation}
and read off each gradient as the factor of its differential.

\noindent\textbf{Step 1: reads (blue).} From
$\mathbf o_m=\bar{\mathbf v}+\mathbf W_2^{\star}\tilde\psi_m$,
$\delta\mathcal L=\sum_m\langle\nabla_{\mathbf o_m},\delta\bar{\mathbf v}+\delta\mathbf W_2^{\star}\,\tilde\psi_m+\mathbf W_2^{\star}\,\delta\tilde\psi_m\rangle$,
so
\begin{equation}
\begin{aligned}
\nabla_{\bar{\mathbf v}}&=\textstyle\sum_m\nabla_{\mathbf o_m},\qquad
\nabla_{\tilde\psi_m}=\mathbf W_2^{\star\top}\nabla_{\mathbf o_m},\\
\nabla_{\mathbf W_2^{\star}}&=\textstyle\sum_m\nabla_{\mathbf o_m}\tilde\psi_m^{\top}.
\end{aligned}
\label{eq:s_b_read}
\end{equation}

\noindent\textbf{Step 2: solve (orange).} Differentiating
$\mathbf W_2^{\star}\mathbf P=\tilde{\mathbf C}+\lambda\mathbf W_2^{\mathrm{init}}$
(\cref{eq:s_solve}) with $\mathbf P=\tilde{\mathbf G}+\lambda\mathbf I$ and
solving for $\delta\mathbf W_2^{\star}$,
\begin{equation}
\begin{aligned}
\delta\mathbf W_2^{\star}=\big[&\delta\tilde{\mathbf C}
+\lambda\,\delta\mathbf W_2^{\mathrm{init}}
-\mathbf W_2^{\star}\,\delta\tilde{\mathbf G}\\
&+\delta\lambda\,(\mathbf W_2^{\mathrm{init}}-\mathbf W_2^{\star})\big]\mathbf P^{-1}.
\end{aligned}
\label{eq:s_b_dW}
\end{equation}
Moving the symmetric $\mathbf P^{-1}$ to the other argument of
$\langle\nabla_{\mathbf W_2^{\star}},\delta\mathbf W_2^{\star}\rangle$
pairs each term of the bracket with
$\mathbf Z:=\nabla_{\mathbf W_2^{\star}}\mathbf P^{-1}$, which gives the
four gradients of the main text:
\begin{equation}
\begin{aligned}
\nabla_{\tilde{\mathbf C}}&=\mathbf Z,&
\nabla_{\tilde{\mathbf G}}&=-\mathbf W_2^{\star\top}\mathbf Z,\\
\nabla_{\mathbf W_2^{\mathrm{init}}}&=\lambda\mathbf Z,&
\nabla_{\lambda}&=\big\langle\mathbf Z,\mathbf W_2^{\mathrm{init}}-\mathbf W_2^{\star}\big\rangle .
\end{aligned}
\label{eq:s_b_solve}
\end{equation}
As $\mathbf P^{-1}=\mathbf L^{-\top}\mathbf L^{-1}$, $\mathbf Z$ takes two
matrix products with the stored $\mathbf L^{-1}$.

\noindent\textbf{Step 3: rescaling (yellow).} The forward relations are
(\cref{sec:s_sums})
\begin{equation}
\begin{aligned}
\tilde{\mathbf G}&=\mathbf S^{-1}\mathbf G'\mathbf S^{-1},\qquad
\tilde{\mathbf C}=\mathbf C'\mathbf S^{-1},\\
\tilde\psi_m&=\mathbf S^{-1}\big(\psi(\hat{\mathbf q}_m)-\bar\psi\big),
\end{aligned}
\label{eq:s_b_fwd3}
\end{equation}
with the feature $\psi(\hat{\mathbf q}_m)$ of a reading token held fixed.
Using
$\delta(\mathbf S^{-1})=-\mathbf S^{-1}\delta\mathbf S\,\mathbf S^{-1}$,
\begin{equation}
\begin{aligned}
\delta\tilde{\mathbf G}&=\mathbf S^{-1}\delta\mathbf G'\mathbf S^{-1}
-\mathbf S^{-1}\delta\mathbf S\,\tilde{\mathbf G}
-\tilde{\mathbf G}\,\delta\mathbf S\,\mathbf S^{-1},\\
\delta\tilde{\mathbf C}&=\delta\mathbf C'\mathbf S^{-1}
-\tilde{\mathbf C}\,\delta\mathbf S\,\mathbf S^{-1},\\
\delta\tilde\psi_m&=-\mathbf S^{-1}\delta\bar\psi
-\mathbf S^{-1}\delta\mathbf S\,\tilde\psi_m .
\end{aligned}
\label{eq:s_b_dresc}
\end{equation}
The quantities of this block, $\mathbf G'$, $\mathbf C'$, $\mathbf S$, and
$\bar\psi$, reach the loss only through $\tilde{\mathbf G}$,
$\tilde{\mathbf C}$, and the $\tilde\psi_m$, so for them
$\delta\mathcal L=\langle\nabla_{\tilde{\mathbf G}},\delta\tilde{\mathbf G}\rangle+\langle\nabla_{\tilde{\mathbf C}},\delta\tilde{\mathbf C}\rangle+\sum_m\langle\nabla_{\tilde\psi_m},\delta\tilde\psi_m\rangle$.
There is no $\delta\bar{\mathbf v}$ term: $\bar{\mathbf v}$ enters
$\mathbf o_m$ directly (\cref{fig:s_chain}), so
$\nabla_{\bar{\mathbf v}}$ is final in \cref{eq:s_b_read}. Inserting
\cref{eq:s_b_dresc}, the terms without $\delta\mathbf S$ give
\begin{equation}
\nabla_{\mathbf C'}=\nabla_{\tilde{\mathbf C}}\mathbf S^{-1},\qquad
\nabla_{\bar\psi}=-\mathbf S^{-1}\textstyle\sum_m\nabla_{\tilde\psi_m},
\label{eq:s_b_c0}
\end{equation}
and contribute
$\mathbf S^{-1}\nabla_{\tilde{\mathbf G}}\mathbf S^{-1}$ to
$\nabla_{\mathbf G'}$. As $\delta\mathbf S$ is diagonal, each of the
four terms with $\delta\mathbf S$ reduces to a sum over $i$, \eg,
$-\langle\nabla_{\tilde{\mathbf G}},\mathbf S^{-1}\delta\mathbf S\,\tilde{\mathbf G}\rangle=-\sum_i\delta S_{ii}\,[\tilde{\mathbf G}\nabla_{\tilde{\mathbf G}}^{\top}]_{ii}/S_{ii}$.
Collecting them, with $\tilde{\mathbf G}$ symmetric,
\begin{equation}
\begin{aligned}
\nabla_{S_{ii}}=-\frac{1}{S_{ii}}\Big(&\big[(\nabla_{\tilde{\mathbf G}}+\nabla_{\tilde{\mathbf G}}^{\top})\tilde{\mathbf G}
+\nabla_{\tilde{\mathbf C}}^{\top}\tilde{\mathbf C}\big]_{ii}\\
&+\textstyle\sum_m(\nabla_{\tilde\psi_m})_i\,\tilde\psi_{m,i}\Big).
\end{aligned}
\label{eq:s_b_S}
\end{equation}
Finally, $S_{ii}^2=(\mathbf G')_{ii}$ (\cref{eq:s_S}) gives
$\delta S_{ii}=\delta(\mathbf G')_{ii}/(2S_{ii})$, which adds a
diagonal term:
\begin{equation}
\nabla_{\mathbf G'}=\mathbf S^{-1}\nabla_{\tilde{\mathbf G}}\mathbf S^{-1}
+\operatorname{diag}\!\Big(\frac{\nabla_{S_{ii}}}{2S_{ii}}\Big).
\label{eq:s_b_c}
\end{equation}

\noindent\textbf{Step 4: tokens (green).} Let
$\psi'_n:=\psi_n-\bar\psi$, $\mathbf v'_n:=\mathbf v_n-\bar{\mathbf v}$ and
$A:=\sum_na_n$, so that
$\mathbf G'=\sum_na_n\psi'_n{\psi'_n}^{\top}$ and
$\mathbf C'=\sum_na_n\mathbf v'_n{\psi'_n}^{\top}$. The means enter
$\delta\mathbf G'$ and $\delta\mathbf C'$ only multiplied by
$\sum_na_n\psi'_n=\mathbf 0$ or $\sum_na_n\mathbf v'_n=\mathbf 0$, so
their gradients pass through \cref{eq:s_means} alone, with
$\delta\bar\psi=\tfrac1A\sum_n(\delta a_n\psi'_n+a_n\delta\psi_n)$
and likewise for $\bar{\mathbf v}$. Collecting the terms of each token,
\begin{equation}
\begin{aligned}
\nabla_{\psi_n}&=a_n\big[(\nabla_{\mathbf G'}+\nabla_{\mathbf G'}^{\top})\psi'_n
+\nabla_{\mathbf C'}^{\top}\mathbf v'_n+\nabla_{\bar\psi}/A\big],\\
\nabla_{\mathbf v_n}&=a_n\big[\nabla_{\mathbf C'}\psi'_n+\nabla_{\bar{\mathbf v}}/A\big],\\
\nabla_{a_n}&={\psi'_n}^{\top}\nabla_{\mathbf G'}\psi'_n
+{\mathbf v'_n}^{\top}\nabla_{\mathbf C'}\psi'_n\\
&\quad+\big(\nabla_{\bar\psi}^{\top}\psi'_n+\nabla_{\bar{\mathbf v}}^{\top}\mathbf v'_n\big)/A .
\end{aligned}
\label{eq:s_b_tokens}
\end{equation}
A reading token receives
$\nabla_{\psi(\hat{\mathbf q}_m)}=\mathbf S^{-1}\nabla_{\tilde\psi_m}$ by
\cref{eq:s_b_fwd3}. From these token gradients, autograd continues to the
projections of the TTS sublayer and to the earlier layers.

\noindent\textbf{Cost.} Stacked over tokens,
\cref{eq:s_b_read,eq:s_b_tokens} are matrix products with token-sized
operands and dominate the cost. All other steps involve only $h\times h$ and
$d\times h$ matrices, whose size does not depend on the number of tokens.
Besides the tokens and the five sums, the backward pass reuses only
$\mathbf W_2^{\star}$ and $\mathbf L^{-1}$ from the forward pass, which
also needs $\mathbf L^{-1}$ for $u_m$
(\cref{eq:s_u_um}). $\mathbf L^{-1}$ is a dense lower-triangular
$h\times h$ matrix, \ie, about 2.4M entries per layer for $h=1536$, for any
number of tokens. Storing it trades this fixed memory for speed: every
product with $\mathbf P^{-1}=\mathbf L^{-\top}\mathbf L^{-1}$ becomes two
dense matrix multiplications, which parallelize well on GPUs, instead of two
triangular solves, and nothing else of the factorization has to be kept, as
it would be for autograd. Computing $\mathbf L^{-1}$ costs $O(h^3)$, the
same order as the Cholesky factorization and small compared with the
$O(Nh^2)$ cost of accumulating the sums once $N\gg h$.

\begin{table*}[t]
\centering
\caption{\textbf{Complete ablation results for the number of input views} on
left-eye Stereo4D~\cite{jin2025stereo4d} at $128\times128$ (256 tokens per
view), from the runs of \cref{tab:pil_arms}. \textbf{Ours} adds the
uncertainty maps (Unc.) to TTS+RoPE and is the configuration of our
large-scale 4D model (\cref{sec:s_train}). Since the uncertainty losses do
not backpropagate into the rendering network, Ours and TTS+RoPE differ only
by training stochasticity. Ours matches or exceeds every ablated variant
and outperforms both baselines by 0.5--2.6\,dB at every $V_{\mathrm{in}}$.
\textsc{FSM}-style writes in chunks of 2{,}048 tokens, \ie, 1 to 16
sequential updates for $V_{\mathrm{in}}=8$ to $128$ (4 during training).
Best in bold, second-best underlined.}
\label{tab:s_scale}
\vspace{-3.0mm}
\setlength{\tabcolsep}{3.6pt}
\resizebox{\linewidth}{!}{
\begin{tabular}{l*{18}{c}}
\toprule
\multirow{2}{*}{Methods} & \multicolumn{3}{c}{$V_{\mathrm{in}}=4$} & \multicolumn{3}{c}{$V_{\mathrm{in}}=8$} & \multicolumn{3}{c}{$V_{\mathrm{in}}=16$} & \multicolumn{3}{c}{$V_{\mathrm{in}}=32$} & \multicolumn{3}{c}{$V_{\mathrm{in}}=64$} & \multicolumn{3}{c}{$V_{\mathrm{in}}=128$} \\
\cmidrule(lr){2-4}\cmidrule(lr){5-7}\cmidrule(lr){8-10}\cmidrule(lr){11-13}\cmidrule(lr){14-16}\cmidrule(lr){17-19}
 & PSNR$\uparrow$ & SSIM$\uparrow$ & LPIPS$\downarrow$ & PSNR$\uparrow$ & SSIM$\uparrow$ & LPIPS$\downarrow$ & PSNR$\uparrow$ & SSIM$\uparrow$ & LPIPS$\downarrow$ & PSNR$\uparrow$ & SSIM$\uparrow$ & LPIPS$\downarrow$ & PSNR$\uparrow$ & SSIM$\uparrow$ & LPIPS$\downarrow$ & PSNR$\uparrow$ & SSIM$\uparrow$ & LPIPS$\downarrow$ \\
\midrule
\textsc{FSM}-style~\cite{fsm}
& 22.19 & 0.646 & 0.229 & 25.96 & 0.788 & 0.120 & 30.43 & 0.900 & 0.054 & 33.63 & 0.948 & 0.029 & 29.50 & 0.890 & 0.073 & 21.89 & 0.651 & 0.290 \\
\textsc{MesaNet}-style~\cite{mesanet}
& 22.88 & 0.651 & 0.243 & 27.15 & 0.811 & 0.115 & 30.12 & 0.892 & 0.059 & 31.78 & 0.925 & 0.039 & 32.33 & 0.932 & 0.035 & 32.19 & 0.930 & 0.035 \\
\midrule
TTS+RoPE
& \textbf{24.11} & \textbf{0.724} & \textbf{0.163} & 28.39 & \textbf{0.856} & \textbf{0.080} & \textbf{31.97} & \textbf{0.924} & \textbf{0.043} & \textbf{34.30} & \textbf{0.953} & \textbf{0.027} & \textbf{35.12} & \textbf{0.961} & \textbf{0.023} & \textbf{34.75} & \underline{0.955} & \textbf{0.025} \\
\quad w/o. RoPE
& 23.56 & 0.697 & 0.196 & 28.26 & \underline{0.855} & \underline{0.083} & 31.65 & \underline{0.923} & \underline{0.044} & 33.81 & 0.951 & \underline{0.028} & 34.47 & 0.957 & 0.025 & \underline{34.39} & \textbf{0.956} & \textbf{0.025} \\
\quad w/o. time in RoPE
& 23.92 & \underline{0.707} & \underline{0.174} & \underline{28.46} & 0.851 & \underline{0.083} & 31.74 & 0.920 & 0.045 & 33.87 & 0.949 & 0.029 & 34.70 & 0.958 & \underline{0.024} & 33.83 & 0.951 & 0.029 \\
\quad w/o. prior $\mathbf W_2^{\mathrm{init}}$
& 23.04 & 0.675 & 0.215 & 27.21 & 0.814 & 0.102 & 31.28 & 0.913 & 0.049 & 33.67 & 0.947 & 0.030 & 34.18 & 0.951 & 0.028 & 34.14 & 0.949 & 0.030 \\
\quad w/o. learnable $a_n$
& 23.33 & 0.684 & 0.207 & 28.22 & 0.844 & 0.091 & 31.78 & 0.920 & 0.046 & 34.06 & 0.951 & \underline{0.028} & 34.79 & 0.958 & \underline{0.024} & 33.43 & 0.948 & 0.032 \\
\midrule
\rowcolor{LightCyan}
\textbf{Ours} (TTS+RoPE+Unc.)
& \underline{23.93} & 0.702 & 0.184 & \textbf{28.50} & 0.847 & 0.090 & \underline{31.90} & 0.922 & 0.045 & \underline{34.11} & \underline{0.952} & \underline{0.028} & \underline{34.95} & \underline{0.960} & \textbf{0.023} & \underline{34.39} & 0.953 & \underline{0.027} \\
\bottomrule
\end{tabular}
}
\vspace{-3mm}
\end{table*}

\subsection{Uncertainty Maps}
\label{sec:s_unc}
We derive the statements of \cref{sec:confidence}: in its paragraph ``Epistemic'',
(i)~$\mathbf W_2^{\star}$ is a MAP estimate for any $\sigma^2$, (ii)~each
row of $\mathbf W_2$ has posterior covariance $\sigma^2\mathbf P^{-1}$,
(iii)~a read has covariance $\sigma^2u_m\mathbf I$ with
$u_m\le\|\tilde\psi_m\|^2/\lambda$, and (iv)~the fusion needs only $L+1$
parameters; in the paragraph ``Training'', (v)~$\mathcal L_{\mathrm{NLL}}$ is a
Gaussian negative log-likelihood.

\noindent\textbf{(i) MAP estimate.} The regression model of
\Cref{sec:confidence} treats the features as fixed and assumes
\begin{equation}
\tilde{\mathbf v}_n=\mathbf W_2\tilde\psi_n+\boldsymbol\varepsilon_n,\qquad
\boldsymbol\varepsilon_n\sim\mathcal N\big(\mathbf 0,\tfrac{\sigma^2}{a_n}\mathbf I\big),
\label{eq:s_u_lik}
\end{equation}
independently over $n$, with a prior in which the entries of $\mathbf W_2$
are independent Gaussians with the entries of $\mathbf W_2^{\mathrm{init}}$
as means and variance $\sigma^2/\lambda$. Up to factors that do not depend
on $\mathbf W_2$, the densities of one value and of the prior are
\begin{equation}
\begin{aligned}
p(\tilde{\mathbf v}_n\mid\mathbf W_2)&\propto
\exp\!\big(-\tfrac{a_n}{2\sigma^2}\|\tilde{\mathbf v}_n-\mathbf W_2\tilde\psi_n\|^2\big),\\
p(\mathbf W_2)&\propto
\exp\!\big(-\tfrac{\lambda}{2\sigma^2}\|\mathbf W_2-\mathbf W_2^{\mathrm{init}}\|_F^2\big).
\end{aligned}
\label{eq:s_u_dens}
\end{equation}
By Bayes' rule~\cite{bishop2006prml}, the posterior is proportional to the
prior times the likelihood, which is the product of the densities of all
values. Multiplying exponentials adds their exponents:
\begin{equation}
\begin{aligned}
&p(\mathbf W_2\mid\{\tilde{\mathbf v}_n\})
\propto p(\mathbf W_2)\textstyle\prod_np(\tilde{\mathbf v}_n\mid\mathbf W_2)\\
&\ \propto\exp\!\Big(-\tfrac{1}{2\sigma^2}\Big[\textstyle\sum_na_n\|\tilde{\mathbf v}_n-\mathbf W_2\tilde\psi_n\|^2\\
&\hspace{6.5em}+\lambda\|\mathbf W_2-\mathbf W_2^{\mathrm{init}}\|_F^2\Big]\Big)\\
&\ =\exp\!\big(-\tfrac{1}{2\sigma^2}\,\mathcal L_{\mathrm{TTS}}(\mathbf W_2)\big),
\end{aligned}
\label{eq:s_u_post}
\end{equation}
since the bracket is the objective \cref{eq:s_obj}.
The posterior is thus largest at the minimizer $\mathbf W_2^{\star}$ of
$\mathcal L_{\mathrm{TTS}}$ (\cref{eq:s_solve}), which is therefore the MAP
estimate for any $\sigma^2$.

\noindent\textbf{(ii) Posterior covariance.} Substituting
$\tilde{\mathbf C}+\lambda\mathbf W_2^{\mathrm{init}}=\mathbf W_2^{\star}\mathbf P$
(\cref{eq:s_solve}) into \cref{eq:s_quad} and completing the square,
\begin{equation}
\begin{aligned}
\mathcal L_{\mathrm{TTS}}(\mathbf W_2)
&=\operatorname{tr}\big(\mathbf W_2\mathbf P\mathbf W_2^{\top}\big)
-2\operatorname{tr}\big(\mathbf W_2\mathbf P\mathbf W_2^{\star\top}\big)+\mathrm{const}\\
&=\operatorname{tr}\big((\mathbf W_2-\mathbf W_2^{\star})\mathbf P(\mathbf W_2-\mathbf W_2^{\star})^{\top}\big)+\mathrm{const}\\
&=\textstyle\sum_{k=1}^{d}(\mathbf w_k-\mathbf w_k^{\star})^{\top}\mathbf P(\mathbf w_k-\mathbf w_k^{\star})+\mathrm{const},
\end{aligned}
\label{eq:s_u_square}
\end{equation}
where $\mathrm{const}$ denotes terms independent of $\mathbf W_2$, and $\mathbf w_k^{\top}$,
$\mathbf w_k^{\star\top}$ are the $k$-th rows of $\mathbf W_2$,
$\mathbf W_2^{\star}$. With \cref{eq:s_u_post},
the posterior factorizes into independent Gaussian rows with mean
$\mathbf w_k^{\star}$ and covariance $\sigma^2\mathbf P^{-1}$.

\noindent\textbf{(iii) Variance of a read.} The $k$-th component of
$\mathbf o_m=\bar{\mathbf v}+\mathbf W_2\tilde\psi_m$ is
$\bar v_k+\tilde\psi_m^{\top}\mathbf w_k$, whose variance is
$\tilde\psi_m^{\top}(\sigma^2\mathbf P^{-1})\tilde\psi_m=\sigma^2u_m$ with
\begin{equation}
u_m=\tilde\psi_m^{\top}\mathbf P^{-1}\tilde\psi_m
=\|\mathbf L^{-1}\tilde\psi_m\|^2 ,
\label{eq:s_u_um}
\end{equation}
using $\mathbf P=\mathbf L\mathbf L^{\top}$. As the rows are independent,
the read has covariance $\sigma^2u_m\mathbf I$. Since
$\tilde{\mathbf G}\succeq0$, the eigenvalues of
$\mathbf P=\tilde{\mathbf G}+\lambda\mathbf I$ are at least $\lambda$, so
$u_m\le\|\tilde\psi_m\|^2/\lambda$, with equality if and only if
$\tilde\psi_m^{\top}\tilde{\mathbf G}\tilde\psi_m=\sum_na_n(\tilde\psi_n^{\top}\tilde\psi_m)^2=0$,
\ie, $\tilde\psi_m$ is orthogonal to all written features. Conversely,
along an eigenvector of $\tilde{\mathbf G}$ with eigenvalue $\gamma$,
$u_m=\|\tilde\psi_m\|^2/(\gamma+\lambda)$, which is small when many written
features align with $\tilde\psi_m$.

\noindent\textbf{(iv) Fusion with $L+1$ parameters.} Layer $l$ yields the
variance $\sigma_l^2u_{m,l}$ with an unknown scale $\sigma_l^2$. We model
the log-variance of the rendering error as a weighted sum of the per-layer
log-variances with weights $\beta_l\ge0$, and split off the running mean
$\bar u_l$ of $u_{m,l}$ in each term:
\begin{equation}
\begin{aligned}
\textstyle\sum_{l=1}^{L}\beta_l\log\big(\sigma_l^2u_{m,l}\big)
={}&\textstyle\sum_{l=1}^{L}\beta_l\log\big(\sigma_l^2\bar u_l\big)\\
&+\textstyle\sum_{l=1}^{L}\beta_l\log\frac{u_{m,l}}{\bar u_l}.
\end{aligned}
\label{eq:s_u_fuse}
\end{equation}
The first sum on the right contains all unknown scales but is the same for
every token, so a single learnable offset $\beta_0$ replaces it. This gives
$\mathcal U^{\mathrm{epi}}_m=\beta_0+\sum_{l}\beta_l\log(u_{m,l}/\bar u_l)$
of \cref{sec:confidence} and is the sense in which $\beta_0$ absorbs $\sigma^2$. The
fusion thus learns only $L+1$ scalars: $\beta_0$ and $\rho_1,\dots,\rho_L$,
with $\beta_l=\operatorname{softplus}(\rho_l)\ge0$.

\noindent\textbf{(v) Loss.} No ground-truth uncertainty exists, so the maps
are trained on the rendering error itself: a good map predicts a large
variance exactly where the error is
large~\cite{kendall2017uncertainties}. Let $e_s^2$ be the squared error at
pixel $s$, averaged over the three color channels and detached. A
log-variance map $\mathcal V$, either $\mathcal U$ or
$\mathcal U^{\mathrm{epi}}$, models the error $e_s$ as a zero-mean Gaussian
with variance $\exp\mathcal V_s$, \ie, with density
\begin{equation}
p(e_s)=\frac{1}{\sqrt{2\pi\exp\mathcal V_s}}\,
\exp\!\Big(-\frac{e_s^2}{2\exp\mathcal V_s}\Big).
\label{eq:s_u_pe}
\end{equation}
This density measures how plausible the observed error is under the
predicted variance: a small variance makes large errors implausible, while
a large variance admits them but lowers the density of every error. A good
map makes the observed errors plausible, \ie, it minimizes
\begin{equation}
-\log p(e_s)
=\tfrac12\big(e_s^2\exp(-\mathcal V_s)+\mathcal V_s\big)+\tfrac12\log2\pi .
\label{eq:s_u_nll1}
\end{equation}
Without the constant, the average over the $HW$ pixels is
\begin{equation}
\mathcal L_{\mathrm{NLL}}(\mathcal V)=\frac1{HW}\sum_{s}\tfrac12\big(e_s^2\exp(-\mathcal V_s)+\mathcal V_s\big).
\label{eq:s_nll}
\end{equation}
The first term penalizes a variance that is too small for the observed
error; the second penalizes a large variance, so the model cannot declare
every pixel uncertain. A summand is minimized at
$\exp\mathcal V_s=e_s^2$, \ie, when the predicted variance equals the
squared error. In
$\mathcal L_{\mathrm{unc}}=\mathcal L_{\mathrm{NLL}}(\mathcal U)+\mathcal L_{\mathrm{NLL}}(\mathcal U^{\mathrm{epi}})$,
the stop-gradient in
$\mathcal U=\operatorname{sg}[\mathcal U^{\mathrm{epi}}]+\mathcal U^{\mathrm{ale}}$
lets the first term train only the aleatoric head and the second only
$\beta_0$ and $\beta_l$. As the squared error $e_s^2$ is detached, neither term changes the
rendering itself.

\section{Ablations: Setup and Complete Results}
\label{sec:s_ablation}

\begin{table}[!b]
\centering
\caption{\textbf{Complete ablation results for resolution and input order} on left-eye
Stereo4D at $V_{\mathrm{in}}=32$. Shuffled: one random input order per clip
at $128\times128$. }
\label{tab:s_arms}
\vspace{-3.0mm}
\setlength{\tabcolsep}{3.6pt}
\resizebox{\linewidth}{!}{
\begin{tabular}{l*{6}{c}}
\toprule
\multirow{2}{*}{Methods} & \multicolumn{3}{c}{$256\times256$} & \multicolumn{3}{c}{Shuffled} \\
\cmidrule(lr){2-4}\cmidrule(lr){5-7}
 & PSNR$\uparrow$ & SSIM$\uparrow$ & LPIPS$\downarrow$ & PSNR$\uparrow$ & SSIM$\uparrow$ & LPIPS$\downarrow$ \\
\midrule
\textsc{FSM}-style~\cite{fsm}
& 18.84 & 0.456 & 0.542 & 30.94 & 0.913 & 0.050 \\
\textsc{MesaNet}-style~\cite{mesanet}
& 21.30 & 0.610 & 0.349 & 31.78 & 0.925 & 0.039 \\
\midrule
TTS+RoPE
& 22.86 & 0.668 & 0.260 & \textbf{34.30} & \textbf{0.953} & \textbf{0.027} \\
\quad w/o. RoPE
& \textbf{23.42} & \textbf{0.691} & \underline{0.244} & 33.81 & 0.951 & \underline{0.028} \\
\quad w/o. time in RoPE
& 22.63 & 0.664 & 0.272 & 33.87 & 0.950 & 0.029 \\
\quad w/o. prior $\mathbf W_2^{\mathrm{init}}$
& 21.70 & 0.611 & 0.305 & 33.67 & 0.947 & 0.030 \\
\quad w/o. learnable $a_n$
& 22.57 & 0.666 & 0.264 & 34.06 & 0.951 & \underline{0.028} \\
\midrule
\rowcolor{LightCyan}
\textbf{Ours} (TTS+RoPE+Unc.)
& \underline{23.26} & \underline{0.689} & \textbf{0.241} & \underline{34.11} & \underline{0.952} & \underline{0.028} \\
\bottomrule
\end{tabular}
}
\end{table}

\noindent\textbf{Setup.} All variants in \cref{sec:pil_setup} use the backbone of \textsc{FSM}~\cite{fsm} ($L=24$, $p=8$, $d=768$, $h=1536$) and are trained from scratch on the
left-eye Stereo4D~\cite{jin2025stereo4d} training videos for 32K steps on 8
NVIDIA H20 GPUs, with a batch size of 16 per GPU. We use AdamW
($\beta=(0.9,0.95)$, weight decay 0.05), a peak learning rate of
$2\times10^{-4}$ with 1{,}000 linear warm-up steps and cosine decay to zero,
gradient clipping at 1.0, and bfloat16 mixed precision. Each sample has 32
input and 32 target views at $128\times128$. 
We take a random 128-frame
window from the first 136 frames of a clip and draw 64 distinct frames from
it, the first 32 as inputs and the rest as targets. Camera centers and times
are normalized as in the recipe of \textsc{FSM}. 
At test time, they are
normalized over the input views only, and the targets share these
transforms. The \textsc{FSM}-style variant uses weight-norm restoration and
elastic consolidation. The \textsc{MesaNet}-style~\cite{mesanet} variant solves on the
uncentered and unscaled sums of \cref{sec:s_sums}, without prior:
$\mathbf W_2^{\star}=\big(\sum_na_n\mathbf v_n\psi_n^{\top}\big)\big(\sum_na_n\psi_n\psi_n^{\top}+\lambda\mathbf I\big)^{-1}$. The training speed is the wall-clock time
of one steady-state step on the 8 GPUs.

\noindent\textbf{Evaluation.} From each of the 5{,}177 left-eye test clips
with at least 136 frames, we fix 8 target frames and draw the inputs at
random from the remaining frames, nested across $V_{\mathrm{in}}$. In the
shuffled setting, we use one random input order per clip.
\Cref{tab:s_scale} completes \cref{tab:pil_arms} with SSIM and LPIPS
for every $V_{\mathrm{in}}$, and \cref{tab:s_arms} with SSIM and LPIPS at
$256\times256$ and for shuffled inputs. 

\section{Large-Scale Training}
\label{sec:s_train}

\subsection{Training Data}
\label{sec:s_data}

\Cref{tab:s_data} lists the 40 dataset sources (see \cref{sec:exp_setup}) with their shares at the end of
training. 38 form the main pool; the two robot-manipulation sources (RH20T and the DROID wrist camera) form a separate pool that supplies 11\% of the
samples of low-resolution steps.

\begin{table}[!b]
\centering
\caption{\textbf{Training datasets}, grouped by content and sorted by share.
Cams: synchronized cameras per time instant; Seq.: training sequences;
Share: percentage of the main-pool samples, with group totals in
parentheses. $^{\ddagger}$Used in the low-resolution pool only.}
\label{tab:s_data}
\vspace{-3.0mm}
\footnotesize
\renewcommand{\arraystretch}{0.95}
\setlength{\tabcolsep}{0pt}
\begin{tabular*}{\linewidth}{@{}l@{\extracolsep{\fill}}r@{\extracolsep{12pt}}rr@{}}
\toprule
Dataset & Cams & Seq. & Share (\%) \\
\midrule
\multicolumn{4}{@{}l}{\textit{Real, dynamic} (19 datasets, 49.2\%)} \\
\addlinespace[1pt]
\quad SpatialVID~\cite{wang2025spatialvid} & 1 & 353{,}636 & 14.0 \\
\quad Stereo4D~\cite{jin2025stereo4d} & 2 & 81{,}293 & 7.0 \\
\quad RealCam-Vid~\cite{realcam} & 1 & 30{,}742 & 4.0 \\
\quad Replay~\cite{shapovalov2023replay} & 11 & 64 & 3.5 \\
\quad Sekai-2~\cite{li2025sekai} & 1 & 8{,}545 & 3.0 \\
\quad OmniWorld CityWalk~\cite{zhou2025omniworld} & 1 & 31{,}365 & 2.5 \\
\quad CoP3D~\cite{sinha2023common} & 1 & 3{,}042 & 2.0 \\
\quad EgoDex~\cite{hoque2025egodex} & 1 & 31{,}777 & 2.0 \\
\quad Waymo~\cite{sun2020waymo} & 3 & 742 & 2.0 \\
\quad DiVA-360~\cite{lu2024diva360} & 41 & 39 & 2.0 \\
\quad CMU Panoptic~\cite{joo2015panoptic} & 8 & 89 & 1.6 \\
\quad OmniWorld EpicKitchen~\cite{zhou2025omniworld} & 1 & 13{,}424 & 1.5 \\
\quad Ego1K~\cite{lee2026ego1k} & 12 & 46 & 1.0 \\
\quad Sekai walk~\cite{li2025sekai} & 1 & 3{,}015 & 0.8 \\
\quad Sekai drone~\cite{li2025sekai} & 1 & 20{,}654 & 0.8 \\
\quad OmniWorld HOI4D~\cite{zhou2025omniworld} & 1 & 1{,}910 & 0.8 \\
\quad Sekai-2 panorama~\cite{li2025sekai} & 1 & 2{,}543 & 0.7 \\
\quad RH20T$^{\ddagger}$~\cite{fang2023rh20t} & 4--9 & 23{,}642 & -- \\
\quad DROID (wrist)$^{\ddagger}$~\cite{khazatsky2024droid} & 1 & 93{,}448 & -- \\
\addlinespace[3pt]
\multicolumn{4}{@{}l}{\textit{Real, static} (7 datasets, 27.5\%)} \\
\addlinespace[1pt]
\quad DL3DV~\cite{ling2024dl3dv} & 1 & 10{,}344 & 12.0 \\
\quad RealEstate10K~\cite{zhou2018stereo} & 1 & 65{,}967 & 4.0 \\
\quad CO3D~\cite{reizenstein2021co3d} & 1 & 32{,}343 & 3.5 \\
\quad uCO3D~\cite{liu2025uco3d} & 1 & 19{,}092 & 2.5 \\
\quad WildRGB-D~\cite{xia2024rgbd} & 1 & 1{,}913 & 2.0 \\
\quad ARKitScenes~\cite{baruch2021arkitscenes} & 1 & 4{,}375 & 2.0 \\
\quad ScanNet++~\cite{yeshwanth2023scannetpp} & 1 & 833 & 1.5 \\
\addlinespace[3pt]
\multicolumn{4}{@{}l}{\textit{Synthetic, dynamic} (10 datasets, 19.7\%)} \\
\addlinespace[1pt]
\quad MultiCamVideo~\cite{bai2025recammaster} & 10 & 13{,}600 & 4.5 \\
\quad Kubric-5D~\cite{anyview} & 16 & 9{,}600 & 4.5 \\
\quad ParallelDomain-4D~\cite{vanhoorick2024gcd} & 19 & 1{,}375 & 4.5 \\
\quad Syn4D~\cite{syn4d} & 8 & 485 & 2.5 \\
\quad OmniWorld Game~\cite{zhou2025omniworld} & 1 & 3{,}501 & 1.0 \\
\quad Kubric~\cite{greff2022kubric} & 1 & 4{,}353 & 1.0 \\
\quad PointOdyssey~\cite{zheng2023pointodyssey} & 1 & 131 & 0.5 \\
\quad Dynamic Replica~\cite{karaev2023dynamicstereo} & 1--2 & 483 & 0.5 \\
\quad Virtual KITTI 2~\cite{cabon2020vkitti2} & 2 & 50 & 0.4 \\
\quad Spring~\cite{mehl2023spring} & 2 & 36 & 0.3 \\
\addlinespace[3pt]
\multicolumn{4}{@{}l}{\textit{Synthetic, static} (4 datasets, 3.4\%)} \\
\addlinespace[1pt]
\quad Hypersim~\cite{roberts2021hypersim} & 1 & 302 & 1.0 \\
\quad TartanAir~\cite{wang2020tartanair} & 1 & 347 & 1.0 \\
\quad BlendedMVS~\cite{yao2020blendedmvs} & 1 & 477 & 1.0 \\
\quad MVS-Synth~\cite{huang2018deepmvs} & 1 & 120 & 0.4 \\
\bottomrule
\end{tabular*}
\end{table}

\begin{table*}[!t]
\centering
\caption{\textbf{4D baselines and our model}: training data, training
resolution, input frames per training sample, and parameters of the released
checkpoints. $^{\ast}$Frozen external network (DINOv2 for \textsc{4DGT},
VGGT for \textsc{FL4N}). $^{\dagger}$Reconstructor only, without the 14B
Wan2.1 video-diffusion refiner, as in the reconstruction experiments of its
paper.}
\label{tab:s_bl}
\vspace{-3.0mm}
\footnotesize
\setlength{\tabcolsep}{6pt}
\begin{tabular}{@{}lp{0.44\linewidth}lll@{}}
\toprule
Methods & Training data & Training resolution & Input frames & Parameters \\
\midrule
\textsc{4DGT}~\cite{4dgt} & \raggedright EgoExo4D, Nymeria, HOT3D, AEA, Epic-Fields, CoP3D, ARKitTrack & $252^2$ to $504^2$ & 16--64 of 128 & 1.22B + 86M$^{\ast}$ \\
\addlinespace[2pt]
\textsc{MoVieS}~\cite{movies} & \raggedright RealEstate10K, TartanAir, MatrixCity, PointOdyssey, Dynamic Replica, Spring, Virtual KITTI~2, Stereo4D & $224^2$ to $518^2$ & 2--13 & 1.04B \\
\addlinespace[2pt]
\textsc{NeoVerse}$^{\dagger}$~\cite{neoverse} & \raggedright PointOdyssey, Dynamic Replica, Kubric, Spring, Virtual KITTI~2, Waymo, TartanAir, BEDLAM, MVS-Synth, GFIE, HOI4D, CoP3D, DL3DV, ScanNet++, ARKitScenes, Hypersim, MapFree & Long side 560 & 2--8 key frames & 1.51B \\
\addlinespace[2pt]
\textsc{FSM}~\cite{fsm} & \raggedright RealEstate10K, DL3DV, PointOdyssey, Spring, Dynamic Replica, MultiCamVideo, Stereo4D & $128^2$ to $256^2$ & 12--64 of 256 & 312M \\
\addlinespace[2pt]
\textsc{C4G}~\cite{c4g} & \raggedright Kubric, EgoExo4D, RealEstate10K, Spring & $224^2$ & 12 & 1.01B \\
\addlinespace[2pt]
\textsc{FL4N}~\cite{fl4n} & \raggedright CamxTime, Kubric4D & $336\times512$ & 8 & 1.70B + 1.13B$^{\ast}$ \\
\midrule
\rowcolor{LightCyan}
\textbf{Ours} & \raggedright 40 datasets (\cref{tab:s_data}) & Short side 128 to 576 & 4--256 & 313M \\
\bottomrule
\end{tabular}
\vspace{-3mm}
\end{table*}

\begin{table}[!b]
\centering
\caption{\textbf{Runtime of the 4D methods} with $V_{\mathrm{in}}=32$ on one
NVIDIA H20 GPU. Enc.: time to process the inputs (s); FPS:
novel views rendered per second afterwards. $^{\dagger}$Reconstructor only.
}
\label{tab:s_fps}
\vspace{-3.0mm}
\setlength{\tabcolsep}{3.6pt}
\resizebox{\linewidth}{!}{
\begin{tabular}{l*{6}{c}}
\toprule
\multirow{2}{*}{Methods} & \multicolumn{2}{c}{DyCheck, $480\times360$} & \multicolumn{2}{c}{NVIDIA, $960\times540$} & \multicolumn{2}{c}{Stereo4D, $512\times512$} \\
\cmidrule(lr){2-3}\cmidrule(lr){4-5}\cmidrule(lr){6-7}
 & Enc.\,(s)$\downarrow$ & FPS$\uparrow$ & Enc.\,(s)$\downarrow$ & FPS$\uparrow$ & Enc.\,(s)$\downarrow$ & FPS$\uparrow$ \\
\midrule
\textsc{4DGT}
& \textbf{0.67} & 46.3 & \underline{3.48} & 14.3 & \textbf{1.14} & 30.8 \\
\textsc{MoVieS}
& 11.41 & 51.5 & 33.57 & 21.3 & 14.30 & 29.3 \\
\textsc{NeoVerse}$^{\dagger}$
& 3.15 & \underline{161} & 15.95 & \textbf{179} & 5.72 & \underline{123} \\
\textsc{FSM}
& 1.69 & 30.3 & 5.62 & 7.6 & 2.44 & 19.6 \\
\textsc{C4G}
& 47.15 & \textbf{333} & 260.76 & \underline{175} & 79.30 & \textbf{270} \\
\textsc{FL4N}
& 3.53 & 4.0 & \textbf{2.22} & 3.0 & 5.92 & 2.3 \\
\midrule
\rowcolor{LightCyan}
\textbf{Ours}
& \underline{1.41} & 34.4 & 4.96 & 8.3 & \underline{2.07} & 22.1 \\
\bottomrule
\end{tabular}
}
\end{table}

\subsection{Sampling Curriculum, Resolution, and Noise}
\label{sec:s_sampling}

All schedules depend on the training progress $\tau\in[0,1]$.

\vspace{0.5mm}
\noindent\textbf{Views and window.} The number of input views is drawn from
seven buckets (4--7, 8--15, 16--31, 32--47, 48--63, 64--128, and above 128),
whose shares move from $(11,19,40,15,10,3.5,1.5)\%$ at the start to
$(7,11,30,20,18,9,5)\%$ at $\tau\ge0.8$. The cap on the number of inputs grows
from 64 to 128 by $\tau=0.7$, and reaches 256 in rare steps. The inputs are sampled from a temporal window of 2.5–24 frames per input (about 10 on average), which widens during training and is capped at the whole sequence. Each step also draws 6–64 target views. For synchronized rigs, 80\% of the windows count time
instants instead of frames. Static sources use the whole sequence, and the
sparse-view sources (CO3D, RealEstate10K, WildRGB-D, and uCO3D) take 4--24
inputs spread over at least 80\% of it.

\vspace{0.5mm}
\noindent\textbf{Resolution.} The short side starts at 128 pixels, and its
cap grows from 384 to 576 pixels by $\tau=0.6$. From $\tau=0.6$ to $1.0$, most
steps use a short side in $[320,576]$. The batch size is the largest that
fits a fixed token budget per GPU, and the largest samples use context
parallelism over up to 16 GPUs.

\vspace{0.5mm}
\noindent\textbf{Camera noise.} The input cameras of a sample are perturbed
with probability 0.7. A level $\eta$, drawn uniformly from $[0,1]$, scales a random
rotation of up to $1.82^\circ\eta$, a translation of $1.29\%\,\eta$ of the
scene extent, and a focal-length change of $1.7\%\,\eta$, so that $\eta=1$
is twice the mean error of \textsc{ViPE}~\cite{huang2025vipe} on the DL3DV
and RealEstate10K videos with ground truth. Target cameras are never
perturbed.

\vspace{0.5mm}
\noindent\textbf{Timestamp noise for static data.} A static scene should
render the same at any time. For static single-camera sources, a sample
keeps its capture times with probability 0.5, has all times set to zero with
probability 0.2, and otherwise draws an independent time per frame,
uniformly over the input span (at least 1\,s). The synthetic static sources
Hypersim, MVS-Synth, and BlendedMVS have no capture times, so they use only
the last two options, with probability 0.5 each. At test time, static scenes
use $t=0$.

\vspace{0.5mm}
\noindent\textbf{Optimization.} We use AdamW with $\beta=(0.9,0.95)$ and
weight decay 0.05. The learning rate warms up over 2{,}500 steps to
$2\times10^{-4}$ and follows a cosine decay over 400K steps to $10^{-5}$. Gradients are clipped at
1.0, and all results use an EMA of the weights (decay 0.9999).
\section{4D Baselines}
\label{sec:s_baselines}
\Cref{tab:s_bl} lists the training setups of the 4D baselines of \cref{sec:exp_protocol}
and of our model, and \cref{tab:s_fps} their runtimes. All methods receive
the same input views, and all except \textsc{C4G}~\cite{c4g} also receive
the ground-truth cameras. For every method, we normalize poses and times
using the input views only, so that the targets do not influence the
inputs. \textsc{C4G} does not accept cameras and estimates its own. To
render the target views, we therefore estimate the target poses with its
test-time photometric alignment, which optimizes each pose to match the
target image. Since this uses the target images, it favors \textsc{C4G}.
For the accompanying video, we apply the alignment estimated on the
evaluation targets to the whole camera path. For \textsc{NeoVerse}, we run
only its reconstructor, without the video-diffusion refiner.

Evaluated with the protocol of its own paper, each baseline matches or
exceeds its reported PSNR (reproduced vs.\ reported): \textsc{FSM} 33.29
vs.\ 32.16 on Stereo4D; \textsc{4DGT} 16.39 vs.\ 16.12, \textsc{FL4N} 19.04
vs.\ 19.02, \textsc{C4G} with test-time alignment 15.30 vs.\ 15.64, and
\textsc{NeoVerse} 11.86 vs.\ 11.56 on DyCheck; \textsc{MoVieS} 20.71 vs.\
19.16 on NVIDIA. Our protocol is harder than theirs in two ways.
(i)~\emph{Resolution}: the baselines train on small square images ($224^2$
for \textsc{C4G}, up to $256^2$ for \textsc{FSM}, $504^2$ for
\textsc{4DGT}, and up to $518^2$ for \textsc{MoVieS}), whereas we test at
native resolution. Tested at its training resolution instead, \textsc{C4G}
gains 6.3\,dB on Stereo4D, but \textsc{FSM} only 0.1--1.5\,dB.
(ii)~\emph{Inputs}: the baselines train on few inputs from short windows
(2--13 frames for \textsc{MoVieS}, up to 64 of 128 frames for
\textsc{4DGT}, and 12--64 of 256 frames for \textsc{FSM}), whereas our test
inputs span the whole sequence. \textsc{4DGT}, for example, drops from
18.62 to 9.43\,dB on DyCheck with 16 inputs instead of 64.

\vspace{0.5mm}
\noindent\textbf{Runtime.} \Cref{tab:s_fps} reports the time to process the
inputs and the rendering speed separately. Our model encodes 32 views of a
DyCheck scene ($480\times360$) in 1.4\,s and of a Stereo4D clip
($512\times512$) in 2.1\,s, second only to \textsc{4DGT} and 12--17\%
faster than \textsc{FSM}, whose architecture is closest to ours.
\textsc{MoVieS} and \textsc{C4G} take 7--53$\times$ longer to encode, and
the reconstructor of \textsc{NeoVerse} alone takes 2--3$\times$ longer.
Once their Gaussians are built, \textsc{C4G} and \textsc{NeoVerse} render
fastest, at over 120 FPS. Our model renders at 8--34 FPS, slightly faster
than \textsc{FSM} and 3--10$\times$ faster than \textsc{FL4N}, whose decoder
attends to the tokens of all inputs for every target. The uncertainty head
leaves the rendering bit-identical and costs about 10\% of the rendering
speed (22.1 vs.\ 24.3 FPS at $512\times512$), for any $V_{\mathrm{in}}$,
because the solved memory has a fixed size.

\begin{figure*}[!t]
\centering
\includegraphics[width=\linewidth]{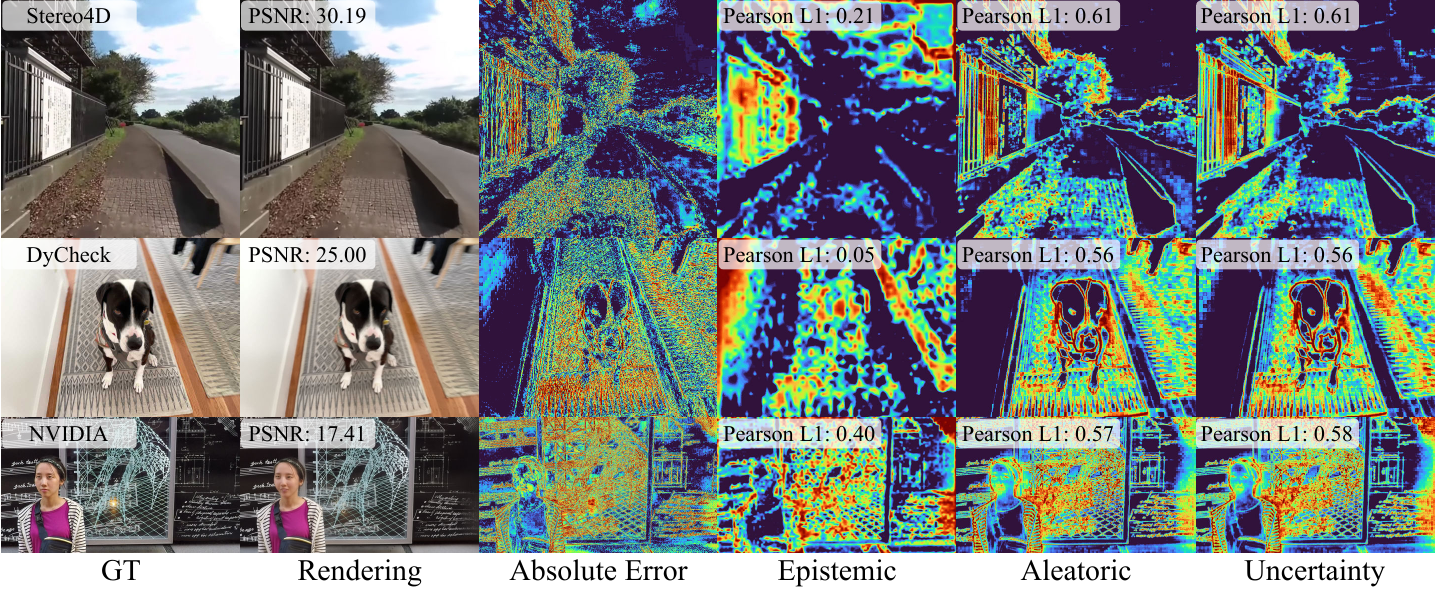}
\vspace{-6mm}
\caption{\textbf{Qualitative evaluations of uncertainty maps on 4D scenes} with $V_{\mathrm{in}}=32$ for different datasets.}
\label{fig:s_unc4d}
\vspace{-10pt}
\end{figure*}

\section{Additional 4D Evaluations}
\label{sec:s_4d}

\subsection{Uncertainty on Dynamic Scenes}
\label{sec:s_unc4d}

\Cref{tab:s_unc4d} reports the Pearson correlation and the AUSE \cite{ilg2018uncertainty,poggi2020uncertainty} of our
uncertainty on the three 4D test sets, on the targets of \cref{tab:4d} and with the protocol and metric code of the static evaluation
(\cref{sec:s_unc3d}). The Pearson $L_1$ lies between 0.55 and 0.59 for every
$V_{\mathrm{in}}$, similar to the static scenes (\cref{tab:s_unc3d}). It
stays stable as $V_{\mathrm{in}}$ grows and the error falls, and rises from
0.55 to 0.58 on DyCheck. The uncertainty is high on moving objects and their
boundaries (the hand in \cref{fig:4d_qual}), on thin structures, and
on surfaces whose appearance changes with the viewpoint, such as reflections
(\cref{fig:s_unc4d}).

\begin{table}[!b]
\centering
\caption{\textbf{Quantitative evaluations of uncertainty on 4D NVS.} Pearson: correlation with the error; AUSE: area under
the sparsification error curve, \ie, how far removing the most uncertain
pixels first is from removing the largest errors first.}
\label{tab:s_unc4d}
\vspace{-3.0mm}
\small
\setlength{\tabcolsep}{0pt}
\begin{tabular*}{\linewidth}{@{\extracolsep{\fill}}llcccc@{}}
\toprule
\multirow{2}{*}{Dataset} & \multirow{2}{*}{$V_{\mathrm{in}}$} & \multicolumn{2}{c}{Pearson$\uparrow$} & \multicolumn{2}{c}{AUSE$\downarrow$} \\
\cmidrule(lr){3-4}\cmidrule(lr){5-6}
 & & $L_1$ & DSSIM & $L_1$ & DSSIM \\
\midrule
\multirow{3}{*}{Stereo4D~\cite{jin2025stereo4d}} & 16 & 0.589 & 0.472 & 0.159 & 0.189 \\
                          & 32 & 0.590 & 0.468 & 0.159 & 0.191 \\
                          & 64 & 0.590 & 0.466 & 0.160 & 0.192 \\
\midrule
\multirow{3}{*}{DyCheck~\cite{gao2022dynamic}}  & 16 & 0.547 & 0.437 & 0.183 & 0.208 \\
                          & 32 & 0.577 & 0.457 & 0.167 & 0.200 \\
                          & 64 & 0.579 & 0.433 & 0.166 & 0.202 \\
\midrule
\multirow{3}{*}{NVIDIA~\cite{yoon2020novel}}   & 16 & 0.568 & 0.470 & 0.177 & 0.199 \\
                          & 32 & 0.566 & 0.454 & 0.178 & 0.201 \\
                          & 64 & 0.564 & 0.455 & 0.180 & 0.202 \\
\bottomrule
\end{tabular*}
\end{table}

\subsection{Time Interpolation and Novel Views}
\label{sec:s_nv}
\begin{figure*}[!t]
\centering
\includegraphics[width=1\linewidth]{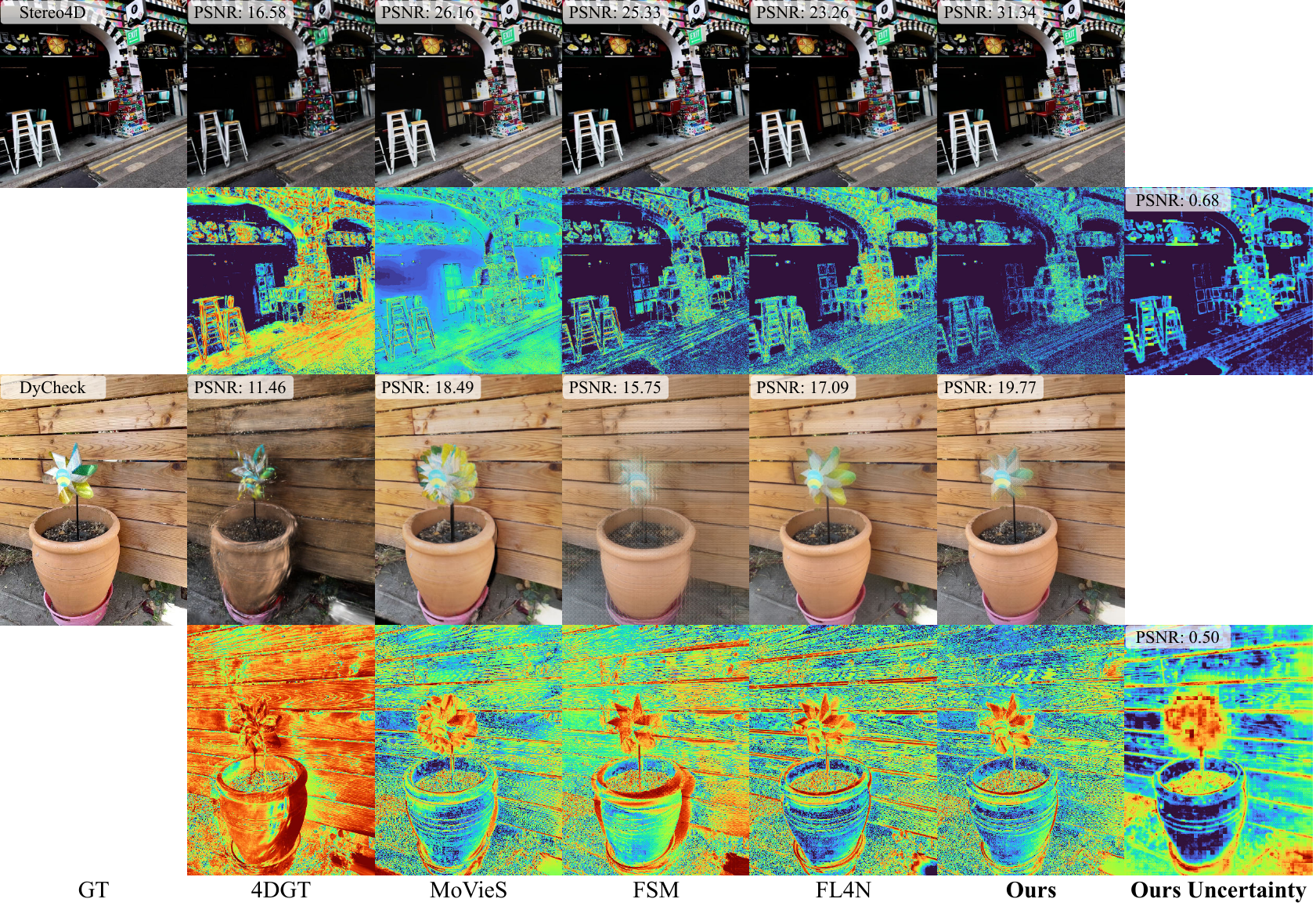}
\vspace{-7mm}
\caption{\textbf{Qualitative comparisons of time interpolation on Stereo4D (top) and novel views on
DyCheck (bottom)} with $V_{\mathrm{in}}=32$. For each dataset, the upper
row shows the ground truth and the renderings with their PSNR, and the lower
row the absolute error of each rendering (log scale; blue: low, red: high).
The last column shows our uncertainty map, rank-matched to our error for
display, with its Pearson $L_1$ correlation.}
\label{fig:s_nv}
\vspace{-2mm}
\end{figure*}

\Cref{sec:exp_protocol} tests each dataset in one setting: on Stereo4D, the right eye
is rendered from left-eye inputs (novel view); on DyCheck, the moving camera
is rendered at times between the inputs (time interpolation).
\Cref{tab:s_nv} swaps the two settings. On Stereo4D, we render the left eye
at 8 held-out time instants per clip (time interpolation). On DyCheck, we
render the two static cameras, which are never given as input (novel view).
Our model is the best in both settings for every $V_{\mathrm{in}}$, by
3.8--6.8\,dB PSNR on Stereo4D. From 32 to 64 input views, \textsc{FSM}
drops by 3.5\,dB, whereas our model keeps improving (\cref{fig:s_nv}).

\begin{table*}[!thb]
\centering
\caption{\textbf{Quantitative comparisons of time interpolation on Stereo4D and novel views on DyCheck.}
Stereo4D: the left eye at 8 held-out instants per clip. DyCheck: the two
static cameras at every validation frame. $^{\dagger}$With test-time pose
alignment.}
\label{tab:s_nv}
\vspace{-3.0mm}
\setlength{\tabcolsep}{3.6pt}
\resizebox{\linewidth}{!}{
\begin{tabular}{l*{18}{c}}
\toprule
\multirow{3}{*}{Methods} & \multicolumn{9}{c}{Stereo4D~\cite{jin2025stereo4d}, time interpolation} & \multicolumn{9}{c}{DyCheck~\cite{gao2022dynamic}, novel views} \\
\cmidrule(lr){2-10}\cmidrule(lr){11-19}
 & \multicolumn{3}{c}{$V_{\mathrm{in}}=16$} & \multicolumn{3}{c}{$V_{\mathrm{in}}=32$} & \multicolumn{3}{c}{$V_{\mathrm{in}}=64$} & \multicolumn{3}{c}{$V_{\mathrm{in}}=16$} & \multicolumn{3}{c}{$V_{\mathrm{in}}=32$} & \multicolumn{3}{c}{$V_{\mathrm{in}}=64$} \\
\cmidrule(lr){2-4}\cmidrule(lr){5-7}\cmidrule(lr){8-10}\cmidrule(lr){11-13}\cmidrule(lr){14-16}\cmidrule(lr){17-19}
 & PSNR$\uparrow$ & SSIM$\uparrow$ & LPIPS$\downarrow$ & PSNR$\uparrow$ & SSIM$\uparrow$ & LPIPS$\downarrow$ & PSNR$\uparrow$ & SSIM$\uparrow$ & LPIPS$\downarrow$ & PSNR$\uparrow$ & SSIM$\uparrow$ & LPIPS$\downarrow$ & PSNR$\uparrow$ & SSIM$\uparrow$ & LPIPS$\downarrow$ & PSNR$\uparrow$ & SSIM$\uparrow$ & LPIPS$\downarrow$ \\
\midrule
\textsc{4DGT}~\cite{4dgt}
& 17.08 & 0.642 & 0.377 & 22.81 & 0.751 & 0.297 & 23.81 & 0.761 & 0.290 & 8.07 & 0.351 & 0.597 & 10.37 & 0.408 & 0.513 & 12.73 & 0.449 & 0.467 \\
\textsc{MoVieS}~\cite{movies}
& 25.88 & \underline{0.836} & 0.156 & 25.51 & 0.838 & 0.153 & 25.02 & \underline{0.833} & \underline{0.157} & 14.38 & 0.498 & 0.405 & 14.37 & 0.505 & 0.418 & 14.30 & 0.505 & 0.422 \\
\textsc{NeoVerse}~\cite{neoverse}
& 16.94 & 0.500 & 0.429 & 17.12 & 0.507 & 0.416 & 17.25 & 0.511 & 0.408 & 11.68 & 0.428 & 0.481 & 11.09 & 0.418 & 0.497 & 10.66 & 0.409 & 0.508 \\
\textsc{FSM}~\cite{fsm}
& \underline{26.64} & 0.829 & \underline{0.153} & \underline{27.86} & \underline{0.857} & \underline{0.135} & 24.41 & 0.740 & 0.225 & 15.01 & 0.486 & 0.410 & 15.47 & 0.505 & 0.378 & 14.58 & 0.456 & 0.498 \\
\textsc{C4G}$^{\dagger}$~\cite{c4g}
& 17.49 & 0.539 & 0.557 & 17.44 & 0.538 & 0.567 & 17.26 & 0.535 & 0.578 & 14.44 & 0.513 & 0.690 & 14.15 & 0.513 & 0.698 & 13.93 & 0.510 & 0.713 \\
\textsc{FL4N}~\cite{fl4n}
& 25.30 & 0.792 & 0.165 & 25.46 & 0.793 & 0.160 & \underline{25.35} & 0.786 & 0.162 & \underline{16.17} & \underline{0.535} & \underline{0.366} & \underline{16.22} & \underline{0.535} & \underline{0.368} & \underline{16.16} & \underline{0.534} & \underline{0.372} \\
\midrule
\rowcolor{LightCyan}
\textbf{Ours}
& \textbf{30.61} & \textbf{0.911} & \textbf{0.081} & \textbf{31.63} & \textbf{0.921} & \textbf{0.072} & \textbf{32.13} & \textbf{0.925} & \textbf{0.068} & \textbf{17.14} & \textbf{0.570} & \textbf{0.267} & \textbf{17.32} & \textbf{0.575} & \textbf{0.269} & \textbf{17.35} & \textbf{0.573} & \textbf{0.269} \\
\bottomrule
\end{tabular}
}
\end{table*}

\subsection{Bullet Time}
\label{sec:s_bullet}

\begin{figure*}[!htb]
\centering
\includegraphics[width=\linewidth]{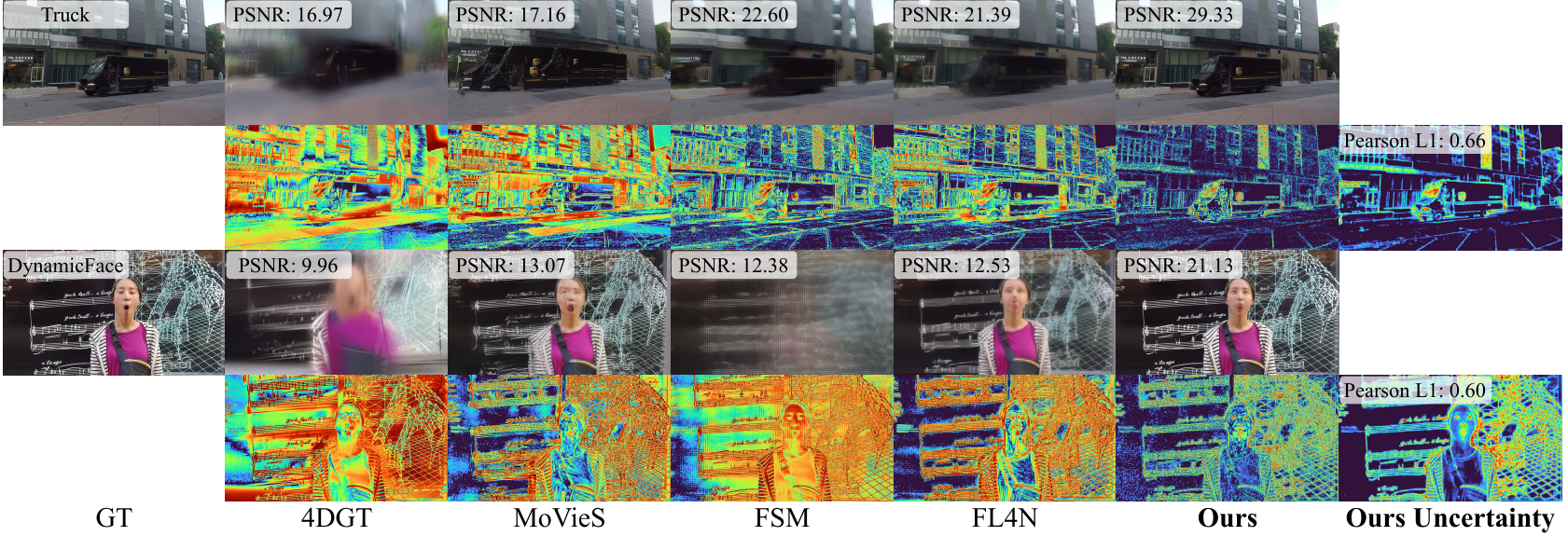}
\vspace{-6mm}
\caption{\textbf{Qualitative comparisons of bullet time on NVIDIA} with $V_{\mathrm{in}}=32$. Top:
Truck; bottom: DynamicFace. Each scene is rendered at a fixed novel time instant with a varying camera.
}
\label{fig:s_bullet}
\vspace{-5mm}
\end{figure*}

We also test bullet time, \ie, freezing the scene at one time instant while
the camera moves around it. Each of the 9 NVIDIA scenes provides 12 cameras
at 12 time instants. For each instant, we render all 12 cameras at that
instant from $V_{\mathrm{in}}=32$ inputs spread uniformly over the other 11
instants and all cameras. No input thus shows the scene at the target time,
so the model must interpolate in time and change the viewpoint at once. The
9 scenes and 12 instants give 108 test cases with 12 target views each,
evaluated as in \cref{tab:4d}. Our model leads by 5.6\,dB PSNR
(\cref{tab:s_bullet}) and is the only method that keeps the moving people
sharp across the viewpoints (\cref{fig:s_bullet}). Its uncertainty reaches
a Pearson $L_1$ of 0.56.

\begin{table}[!b]
\centering
\vspace{-4mm}
\caption{\textbf{Quantitative comparisons of bullet time on NVIDIA} with $V_{\mathrm{in}}=32$: all 12
cameras are rendered at a held-out time instant (108 test cases).}
\label{tab:s_bullet}
\vspace{-3.0mm}
\resizebox{0.75\linewidth}{!}{
\begin{tabular}{lccc}
\toprule
Methods & PSNR$\uparrow$ & SSIM$\uparrow$ & LPIPS$\downarrow$ \\
\midrule
\textsc{4DGT}~\cite{4dgt}           & 14.24 & 0.366 & 0.729 \\
\textsc{MoVieS}~\cite{movies}       & 16.98 & \underline{0.415} & \underline{0.396} \\
\textsc{NeoVerse}~\cite{neoverse}   & 12.69 & 0.222 & 0.605 \\
\textsc{FSM}~\cite{fsm}             & \underline{18.80} & 0.362 & 0.449 \\
\textsc{C4G}$^{\dagger}$~\cite{c4g} & 15.22 & 0.376 & 0.915 \\
\textsc{FL4N}~\cite{fl4n}           & 17.93 & 0.375 & 0.452 \\
\midrule
\rowcolor{LightCyan}
\textbf{Ours}                       & \textbf{24.44} & \textbf{0.746} & \textbf{0.126} \\
\bottomrule
\end{tabular}
}
\end{table}

\subsection{Resolution}
\label{sec:s_res4d}

\begin{figure}[!b]
\centering
\includegraphics[width=\linewidth]{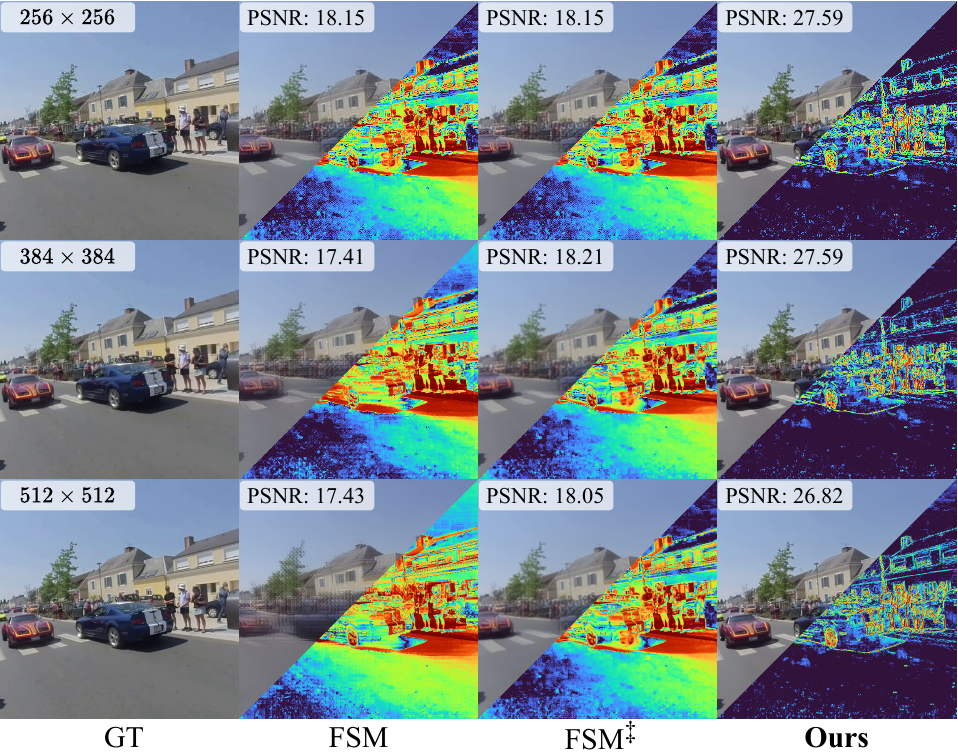}
\vspace{-5pt}
\caption{\textbf{Qualitative comparisons across resolutions on Stereo4D} with $V_{\mathrm{in}}=16$. We compare our model
with \textsc{FSM} run at the test resolution and with
\textsc{FSM}$^{\ddagger}$ run at $256\times256$ and upsampled. Each cell
shows the rendering and its absolute error.}
\label{fig:s_res4d}
\end{figure}

\cref{fig:teaser} illustrates that for the clip shown our quality holds as the
resolution grows. \Cref{tab:s_res4d} confirms this on the Stereo4D test
clips with $V_{\mathrm{in}}=16$. We resize inputs and targets from the
native $512\times512$ to $256\times256$ and $384\times384$, scale the
intrinsics accordingly, and evaluate at the resized resolution. As
\textsc{FSM} is trained at $256\times256$, we run it in two ways: directly
at the test resolution and at $256\times256$ with its rendering upsampled
bilinearly. The PSNR of our model varies by less than 1\,dB across the
three resolutions and is the highest at each: it exceeds the strongest
baseline, \textsc{MoVieS}, by 3.6--4.3\,dB and both \textsc{FSM} variants by
more than 6\,dB (\cref{fig:s_res4d}).

\begin{table}[!b]
\centering
\vspace{-2mm}
\caption{\textbf{Quantitative comparison across resolutions on Stereo4D}
(right eye, $V_{\mathrm{in}}=16$). Inputs and targets are resized from the
native $512\times512$ to each test resolution, at which all methods are
evaluated. \textsc{FSM}$^{\ddagger}$: \textsc{FSM} run at its training
resolution $256\times256$, with its renderings upsampled bilinearly. }
\label{tab:s_res4d}
\vspace{-3.0mm}
\setlength{\tabcolsep}{3.6pt}
\resizebox{\linewidth}{!}{
\begin{tabular}{l*{9}{c}}
\toprule
\multirow{2}{*}{Methods} & \multicolumn{3}{c}{PSNR$\uparrow$} & \multicolumn{3}{c}{SSIM$\uparrow$} & \multicolumn{3}{c}{LPIPS$\downarrow$} \\
\cmidrule(lr){2-4}\cmidrule(lr){5-7}\cmidrule(lr){8-10}
 & $256^2$ & $384^2$ & $512^2$ & $256^2$ & $384^2$ & $512^2$ & $256^2$ & $384^2$ & $512^2$ \\
\midrule
\textsc{4DGT}                & 11.26 & 14.06 & 16.27 & 0.326 & 0.499 & 0.567 & 0.489 & 0.433 & 0.418 \\
\textsc{MoVieS}              & \underline{21.91} & \underline{22.65} & \underline{22.26} & \underline{0.685} & \underline{0.715} & \underline{0.706} & \underline{0.225} & \underline{0.219} & \underline{0.240} \\
\textsc{NeoVerse}            & 17.02 & 16.71 & 16.15 & 0.459 & 0.473 & 0.469 & 0.388 & 0.428 & 0.461 \\
\textsc{FSM}                 & 19.57 & 19.71 & 19.23 & 0.561 & 0.588 & 0.585 & 0.327 & 0.352 & 0.395 \\
\textsc{FSM}$^{\ddagger}$    & 19.57 & 19.90 & 19.31 & 0.561 & 0.599 & 0.595 & 0.327 & 0.376 & 0.416 \\
\textsc{FL4N}                & 20.20 & 20.98 & 21.43 & 0.545 & 0.615 & 0.656 & 0.432 & 0.304 & 0.300 \\
\midrule
\rowcolor{LightCyan}
\textbf{Ours}                & \textbf{26.21} & \textbf{26.62} & \textbf{25.81} & \textbf{0.811} & \textbf{0.820} & \textbf{0.803} & \textbf{0.109} & \textbf{0.129} & \textbf{0.155} \\
\bottomrule
\end{tabular}
}
\end{table}

\begin{figure*}[!t]
\centering
\includegraphics[width=\linewidth]{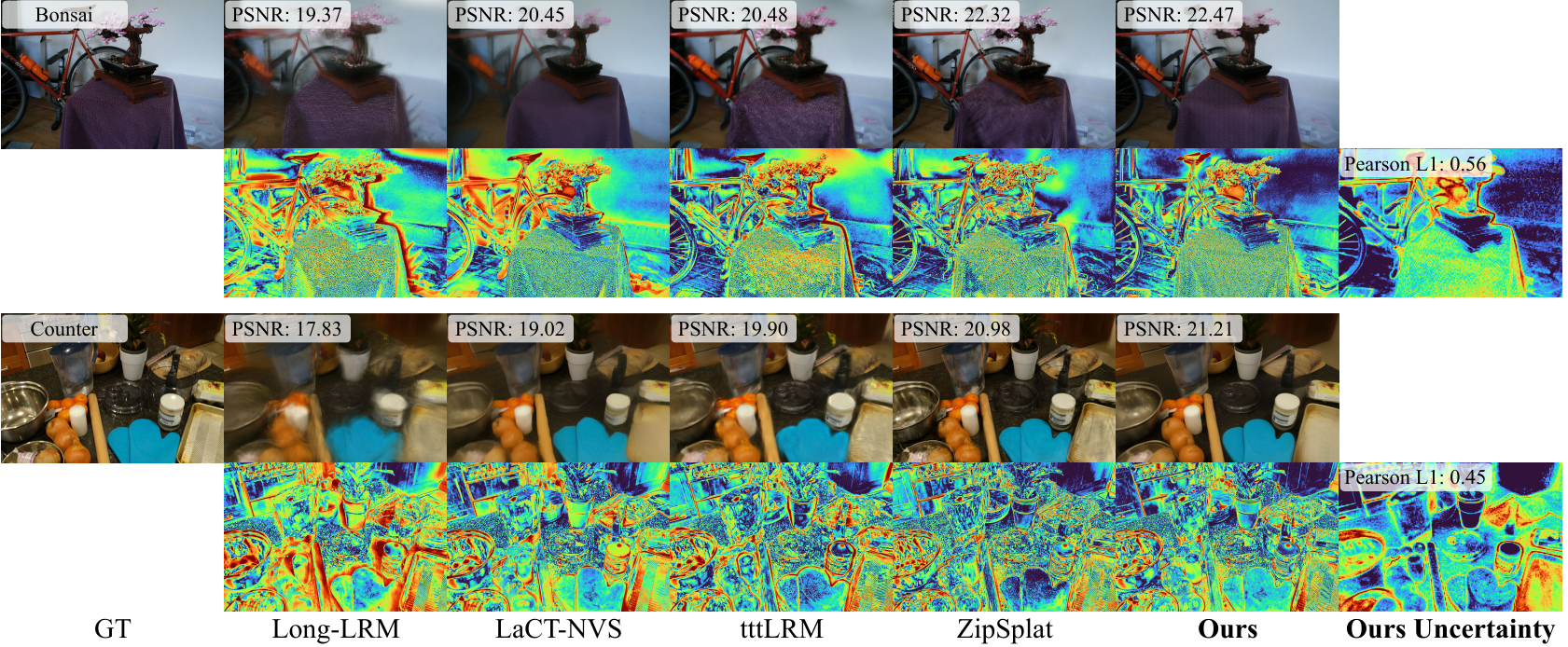}
\vspace{-7mm}
\caption{\textbf{Qualitative 3D NVS comparisons on Mip-NeRF 360} with
$V_{\mathrm{in}}=16$ at the 3DGS evaluation
resolution.
}
\label{fig:s_3d}
\vspace{-12pt}
\end{figure*}

\section{Static 3D NVS and Uncertainty}
\label{sec:s_3d}
A static scene is the special case of a 4D scene in which nothing moves.
As stated in \cref{sec:exp_protocol}, we test our 4D model on this case
in two ways. First, we compare its renderings with those of
state-of-the-art feed-forward 3D NVS models trained only on static scenes;
our model remains competitive although it is trained mainly on dynamic
scenes. Second, we compare its uncertainty maps with uncertainty estimates
obtained by per-scene optimization (\cref{sec:s_unc3d}); our model is more
accurate although it predicts the uncertainty in a single forward pass.

\noindent\textbf{Evaluation data.} We evaluate zero-shot on Mip-NeRF
360~\cite{barron2022mip}, on which no method is trained, at the 3DGS
evaluation resolution: outdoor and indoor images are downsampled $4\times$
and $2\times$ (up to $1558\times1038$). Every 8th image is a target, and the
$V_{\mathrm{in}}$ inputs are spread uniformly over the remaining images. We
also report DL3DV-140~\cite{ling2024dl3dv} at 960p (\cref{tab:s_dl3dv}), on
its 133 usable scenes, with inputs selected by K-means and every 8th frame
as a target. Several methods, including ours, train on part of the DL3DV
training split.

\noindent\textbf{Baselines.} We compare with eight recent feed-forward 3D
NVS methods trained on static scenes. Three predict 3D Gaussians from posed
images: \textsc{Long-LRM}~\cite{longlrm} with a hybrid Mamba2--Transformer,
\textsc{MVP}~\cite{kang2026multi} with a multi-view pyramid Transformer, and
\textsc{AdaptSplat}~\cite{xing2026adaptsplat} with an adapted DINOv3
backbone. Two need no input poses: \textsc{ZipSplat}~\cite{veicht2026zipsplat}
predicts a compact set of Gaussians not tied to image pixels, while
\textsc{LVSPM}~\cite{chen2026lvspm} estimates the poses and renders target
views directly. The remaining three store the inputs in fast weights, as our
model does: \textsc{tttLRM}~\cite{wang2026tttlrm} decodes them into
Gaussians, while \textsc{LaCT-NVS}~\cite{zhang2026test} and
\textsc{FSM}-3D render target views directly. \textsc{FSM}-3D is
\textsc{FSM}~\cite{fsm} trained by its authors on static scenes only. We run
all methods from their released weights, with the same inputs and evaluation
protocol.

\noindent\textbf{Analysis.} Although trained for 4D NVS, our model is
competitive with these specialized 3D models (\cref{tab:s_mip}). It has the
highest PSNR at 16 views, is second to \textsc{LaCT-NVS} at 32 and 64 views,
and has by far the lowest LPIPS, although static scenes make up less than
31\% of its training samples (\cref{tab:s_data}) and its memory is half as
wide as that of \textsc{LaCT-NVS} and \textsc{tttLRM} ($h=1536$ vs.\ 3072).
As our memory accumulates fixed-size sums in chunks, it runs on one GPU at
64 views and $1558\times1038$, whereas \textsc{tttLRM} and
\textsc{LaCT-NVS} keep every token for their updates and need 2--4 GPUs.
Some baselines score below their published numbers, which were obtained near
their $960\times540$ training resolution. They thus generalize less well to
higher resolutions than our model, although our training steps use short
sides of only 128 to 576 pixels (\cref{sec:s_sampling}). On DL3DV-140, our
model also surpasses several pure 3D models (\cref{tab:s_dl3dv}):
\textsc{FSM}-3D, \textsc{ZipSplat}, and \textsc{LVSPM} for every
$V_{\mathrm{in}}$, and \textsc{Long-LRM} at 16 and 64 views.
\Cref{fig:s_3d} shows renderings, errors, and our uncertainty.

\begin{table}[!b]
\centering
\caption{\textbf{Quantitative 3D NVS comparison on Mip-NeRF 360} at the 3DGS
evaluation resolution. $^{\dagger}$Pose-free; $\mathrm{Sim}(3)$-aligned on
the input cameras. $^{*}$Exceeds one GPU at $V_{\mathrm{in}}=64$; run with
equivalent context parallelism or chunking.}
\label{tab:s_mip}
\vspace{-3.0mm}
\setlength{\tabcolsep}{3.2pt}
\resizebox{\linewidth}{!}{
\begin{tabular}{l*{9}{c}}
\toprule
Methods & \multicolumn{3}{c}{PSNR$\uparrow$} & \multicolumn{3}{c}{SSIM$\uparrow$} & \multicolumn{3}{c}{LPIPS$\downarrow$} \\
\cmidrule(lr){2-4}\cmidrule(lr){5-7}\cmidrule(lr){8-10}
$V_{\mathrm{in}}$ & 16 & 32 & 64 & 16 & 32 & 64 & 16 & 32 & 64 \\
\midrule
\textsc{Long-LRM}$^{*}$~\cite{longlrm}        & 18.27 & 19.55 & 18.82 & 0.442 & 0.470 & 0.457 & 0.521 & 0.480 & 0.502 \\
\textsc{LaCT-NVS}$^{*}$~\cite{zhang2026test}        & \underline{19.84} & \textbf{21.51} & \textbf{22.30} & \underline{0.474} & \underline{0.509} & \underline{0.523} & \underline{0.448} & \underline{0.404} & \underline{0.390} \\
\textsc{MVP}~\cite{kang2026multi}                   & 17.33 & 18.32 & 19.12 & 0.426 & 0.449 & 0.467 & 0.554 & 0.516 & 0.482 \\
\textsc{tttLRM}$^{*}$~\cite{wang2026tttlrm}          & 19.72 & 20.57 & 21.11 & \textbf{0.498} & \textbf{0.529} & \textbf{0.552} & 0.456 & 0.417 & 0.397 \\
\textsc{FSM}-3D~\cite{fsm}                & 18.26 & 19.04 & 18.84 & 0.360 & 0.379 & 0.387 & 0.503 & 0.492 & 0.515 \\
\textsc{AdaptSplat}$^{*}$~\cite{xing2026adaptsplat}      & 18.19 & 19.08 & 19.54 & 0.454 & 0.476 & 0.489 & 0.508 & 0.468 & 0.441 \\
\textsc{ZipSplat}$^{\dagger}$~\cite{veicht2026zipsplat}  & 19.79 & 20.77 & 21.16 & 0.460 & 0.489 & 0.503 & 0.460 & 0.423 & 0.401 \\
\textsc{LVSPM}$^{\dagger}$~\cite{chen2026lvspm}     & 17.96 & 18.50 & 19.02 & 0.422 & 0.429 & 0.435 & 0.576 & 0.562 & 0.549 \\
\midrule
\rowcolor{LightCyan}
\textbf{Ours}                  & \textbf{20.17} & \underline{20.93} & \underline{21.33} & 0.451 & 0.467 & 0.475 & \textbf{0.353} & \textbf{0.333} & \textbf{0.323} \\
\bottomrule
\end{tabular}
}
\vspace{-3mm}
\end{table}

\begin{table}[!b]
\centering
\vspace{-3mm}
\caption{\textbf{Quantitative 3D NVS comparison on
DL3DV-140}~\cite{ling2024dl3dv} at 960p. $^{\dagger}$Pose-free;
$\mathrm{Sim}(3)$-aligned on the input cameras. Best of each group in
bold.}
\label{tab:s_dl3dv}
\vspace{-3.0mm}
\setlength{\tabcolsep}{3.2pt}
\resizebox{\linewidth}{!}{
\begin{tabular}{l*{9}{c}}
\toprule
Methods & \multicolumn{3}{c}{PSNR$\uparrow$} & \multicolumn{3}{c}{SSIM$\uparrow$} & \multicolumn{3}{c}{LPIPS$\downarrow$} \\
\cmidrule(lr){2-4}\cmidrule(lr){5-7}\cmidrule(lr){8-10}
$V_{\mathrm{in}}$ & 16 & 32 & 64 & 16 & 32 & 64 & 16 & 32 & 64 \\
\midrule
\multicolumn{10}{l}{\textit{3D models}} \\
\textsc{Long-LRM}~\cite{longlrm}              & 21.01 & 23.92 & 23.56 & 0.721 & 0.784 & 0.795 & 0.329 & 0.269 & 0.263 \\
\textsc{LaCT-NVS}~\cite{zhang2026test}              & \textbf{24.68} & \textbf{26.87} & \textbf{28.30} & 0.794 & 0.837 & 0.857 & \textbf{0.224} & 0.185 & 0.169 \\
\textsc{AdaptSplat}~\cite{xing2026adaptsplat}            & 23.97 & 26.18 & 27.90 & \textbf{0.812} & \textbf{0.858} & \textbf{0.890} & 0.232 & \textbf{0.180} & \textbf{0.146} \\
\textsc{MVP}~\cite{kang2026multi}                   & 23.68 & 25.88 & 27.66 & 0.798 & 0.847 & 0.881 & 0.242 & 0.189 & 0.155 \\
\textsc{tttLRM}~\cite{wang2026tttlrm}                & 23.53 & 25.01 & 25.89 & 0.784 & 0.822 & 0.844 & 0.256 & 0.218 & 0.198 \\
\textsc{FSM}-3D~\cite{fsm}                & 19.74 & 20.62 & 20.23 & 0.554 & 0.585 & 0.566 & 0.431 & 0.402 & 0.455 \\
\textsc{ZipSplat}$^{\dagger}$~\cite{veicht2026zipsplat}  & 20.86 & 21.74 & 22.19 & 0.661 & 0.683 & 0.693 & 0.338 & 0.308 & 0.289 \\
\textsc{LVSPM}$^{\dagger}$~\cite{chen2026lvspm}     & 20.00 & 20.65 & 21.15 & 0.612 & 0.624 & 0.636 & 0.431 & 0.411 & 0.399 \\
\midrule
\multicolumn{10}{l}{\textit{4D models}} \\
\textsc{FSM}~\cite{fsm}                   & 19.49 & 20.05 & 17.38 & 0.552 & 0.566 & 0.474 & 0.434 & 0.420 & 0.513 \\
\rowcolor{LightCyan}
\textbf{Ours}                  & \textbf{22.37} & \textbf{23.34} & \textbf{23.92} & \textbf{0.704} & \textbf{0.727} & \textbf{0.738} & \textbf{0.232} & \textbf{0.210} & \textbf{0.198} \\
\bottomrule
\end{tabular}
}
\vspace{-3mm}
\end{table}

\subsection{Uncertainty on Static Scenes}
\label{sec:s_unc3d}

\noindent\textbf{Setup.} We follow the protocol and metric code of
\textsc{GS-U}~\cite{galappaththige2026predictive} on Tanks and Temples and
Mip-NeRF 360 where every 8th image is a test view and all others are inputs, at
the 3DGS evaluation resolution. For each test image, we compute the area
under the sparsification error curve (AUSE) and the Pearson correlation
between the predicted uncertainty and the per-pixel $L_1$ or DSSIM error,
and average both over images and scenes. Our uncertainty is the predicted
standard deviation $\exp(\mathcal U/2)$. Both metrics depend only on how
well the uncertainty ranks or follows the error, not on its scale, so
methods with different rendering quality can be compared.

\noindent\textbf{Baselines.} \textsc{FisherRF}~\cite{jiang2024fisherrf},
\textsc{Manifold}~\cite{lyu2024manifold}, \textsc{Var3DGS}~\cite{var3dgs},
and \textsc{GS-U} estimate the uncertainty of a 3D Gaussian scene optimized
on each test scene. We reran all four; our runs reproduce the published
\textsc{GS-U} table within 0.05, and within 0.02 for most entries.

\begin{figure*}[!h]
\centering
\includegraphics[width=\linewidth]{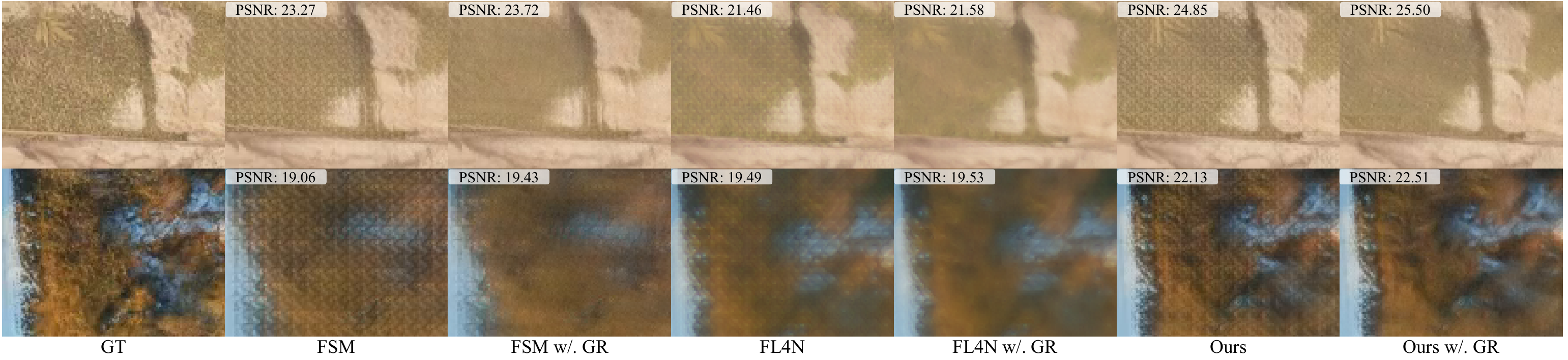}
\vspace{-7mm}
\caption{\textbf{Grid removal.} Renderings before and after removing the
patch-spaced grid.}
\vspace{-12pt}
\label{fig:s_grid}
\end{figure*}

\noindent\textbf{Analysis.} Our model has the best (\ie lowest) AUSE for both error
types and the highest Pearson $L_1$ on both datasets (\cref{tab:s_unc3d}).
On Mip-NeRF 360, for example, its AUSE $L_1$ is 0.213 vs.\ 0.328 for
\textsc{GS-U}, and its Pearson $L_1$ is 0.519 vs.\ 0.368. \textsc{GS-U}
correlates better with the DSSIM error, which measures local structure
instead of per-pixel color, likely because our uncertainty is trained on the
per-pixel squared error (\cref{eq:s_nll}). Unlike the baselines, our model
needs no per-scene optimization: it predicts the uncertainty together with
the rendering in one forward pass, whereas \textsc{GS-U} takes 1.5 to 2.7
hours per scene in our runs. 

\begin{figure}[!b]
\centering
\vspace{-5mm}
\includegraphics[width=\linewidth]{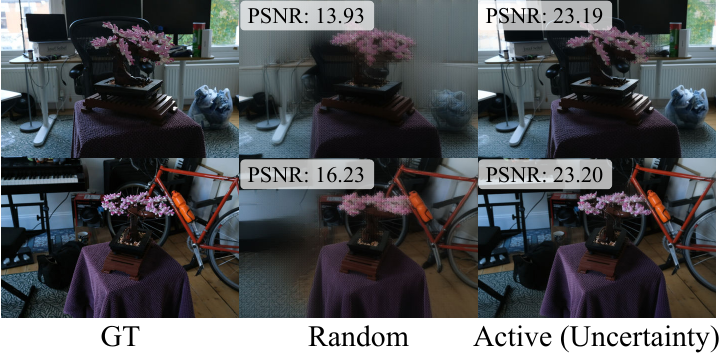}
\vspace{-7mm}
\caption{\textbf{Active view selection with our uncertainty} on the
\textsc{bonsai} scene of Mip-NeRF~360. Two held-out views (rows) rendered
from 8 input views selected at random or by our uncertainty.}
\label{fig:s_active}
\end{figure}

\section{Application: Active View Selection}
\label{sec:s_active}
Our uncertainty maps can be used to decide which view to capture next. Existing active
view selection methods for radiance fields retrain a per-scene model after
every selection and score candidates by the Fisher information of its
parameters~\cite{jiang2024fisherrf,pan2022activenerf} or with a stochastic
variant of 3DGS~\cite{lyu2024manifold,galappaththige2026predictive}. Our
model needs no retraining: it writes the views captured so far into its
memory, renders all candidate views with their uncertainty maps in one
batched forward pass, and selects the most uncertain candidate, \ie, the one
with the largest mean predicted standard deviation.

\noindent\textbf{Setup.} Following FisherRF~\cite{jiang2024fisherrf}, we use
the \textsc{bonsai} scene of Mip-NeRF~360~\cite{barron2022mip}. The 255 views
of the 3DGS training split are candidates, and the other 37 are held out.
Starting from the same four views with maximal pairwise camera distance, we
add one view per step, selected either by our predicted uncertainty (`Active') or
uniformly at random for comparison.

\noindent\textbf{Results.} With eight input views, active selection renders the
held-out views far more accurately than random selection
(\cref{fig:s_active}). The most uncertain candidate usually shows a part of
the scene that no input covers yet, so active selection quickly spreads the
inputs around the scene, whereas random selection leaves regions unobserved, resulting in blurring. As each step costs one forward pass, thus could also
run online, \eg, to guide a robot toward the views where the model is most
uncertain, as a next-best-view policy.

\begin{table}[!b]
\centering
\vspace{-4mm}
\caption{\textbf{Quantitative comparison of uncertainty estimation on 3D
NVS}, with the protocol of
\textsc{GS-U}~\cite{galappaththige2026predictive}. AUSE~\cite{ilg2018uncertainty,poggi2020uncertainty}
and Pearson correlation are computed against the $L_1$ and DSSIM errors.
Best in bold, second-best underlined.}
\vspace{-2mm}
\label{tab:s_unc3d}

\setlength{\tabcolsep}{3.6pt}
\resizebox{\linewidth}{!}{
\begin{tabular}{l*{8}{c}}
\toprule
\multirow{3}{*}{Methods} & \multicolumn{4}{c}{Tanks and Temples~\cite{knapitsch2017tanks}} & \multicolumn{4}{c}{Mip-NeRF 360~\cite{barron2022mip}} \\
\cmidrule(lr){2-5}\cmidrule(lr){6-9}
 & \multicolumn{2}{c}{AUSE$\downarrow$} & \multicolumn{2}{c}{Pearson$\uparrow$} & \multicolumn{2}{c}{AUSE$\downarrow$} & \multicolumn{2}{c}{Pearson$\uparrow$} \\
\cmidrule(lr){2-3}\cmidrule(lr){4-5}\cmidrule(lr){6-7}\cmidrule(lr){8-9}
 & $L_1$ & DSSIM & $L_1$ & DSSIM & $L_1$ & DSSIM & $L_1$ & DSSIM \\
\midrule
\textsc{FisherRF}~\cite{jiang2024fisherrf}
& 0.712 & 0.708 & $-$0.087 & $-$0.119 & 0.710 & 0.607 & $-$0.087 & 0.003 \\
\textsc{Manifold}~\cite{lyu2024manifold}
& 0.555 & 0.631 & 0.051 & 0.009 & 0.518 & 0.551 & 0.073 & $-$0.002 \\
\textsc{Var3DGS}~\cite{var3dgs}
& 0.529 & 0.522 & 0.152 & 0.194 & 0.559 & 0.480 & 0.111 & 0.165 \\
\textsc{GS-U}~\cite{galappaththige2026predictive}
& \underline{0.311} & \underline{0.217} & \underline{0.423} & \textbf{0.586} & \underline{0.328} & \underline{0.216} & \underline{0.368} & \textbf{0.546} \\
\midrule
\rowcolor{LightCyan}
\textbf{Ours}
& \textbf{0.177} & \textbf{0.158} & \textbf{0.569} & \underline{0.564} & \textbf{0.213} & \textbf{0.165} & \textbf{0.519} & \underline{0.455} \\
\bottomrule
\end{tabular}
}
\end{table}

\section{Grid Artifacts}
\label{sec:s_grid}

Our rendered views, like \textsc{FSM}~\cite{fsm},  exhibit faint grid artifacts in smooth areas such as foliage, snow, and walls
(\cref{fig:s_grid})~\cite{odena2016deconvolution}. This  comes from the
decoder, which maps each token to its own $p\times p$ pixel block with a
shared linear layer and no overlap between blocks, so similar neighboring
tokens can produce the same small pattern in every block. The pattern is easy to
 but accounts for less than 1\% of the squared error for us, and a similar amount for \textsc{FSM}, and thus yields
almost no gradient. A better memory therefore does not remove it.

We remove the grid after rendering, without training, by exploiting the fact
that it repeats from patch to patch, whereas content rarely does.
Subtracting from each pixel its mean over the surrounding $p\times p$ window
removes smooth content but keeps the grid, which has zero mean over any such
window. For each of the $p^2$ offsets within a patch, we average this
residual over the surrounding $5\times5$ patches, weighting flat patches
more since they show the grid without content, and subtract the resulting
zero-mean pattern, which leaves the brightness of the patch unchanged. On
267 held-out views of 28 scenes ($320\times480$, 32 inputs), this single
pass lowers the strongest spectral peak of the grid from $12.9\times$ to
$1.7\times$ the background level ($0.95\times$ for the ground truth). The filter applies to any method that decodes each token into its own pixel block, \eg, \textsc{FSM} and \textsc{FL4N}. 
Note, we do not apply this grid artifact removal step for any for the quantitative results presented in the paper.

\end{document}